\documentclass[sigconf]{acmart}

\usepackage[utf8]{inputenc} 
\usepackage[T1]{fontenc}    
\usepackage{hyperref}       
\usepackage{url}            
\usepackage{booktabs}       
\usepackage{amsfonts}       
\usepackage{nicefrac}       
\usepackage{microtype}      

\usepackage{wrapfig,lipsum,booktabs}
\usepackage{multirow}

\usepackage{hyperref}

\usepackage{math_commands}

\usepackage{hyperref}
\usepackage{url}
\usepackage{multirow}
\usepackage{graphicx}
\usepackage{booktabs}
\usepackage{caption}
\usepackage{graphicx}
\usepackage{subfigure}

\usepackage{amsthm}

\usepackage{enumitem}
\usepackage{bm}
\newcommand{\eat}[1]{}
\newtheorem{theorem}{Theorem}

\theoremstyle{definition}

\usepackage{xspace}
\mathchardef\mhyphen="2D
\newcommand{\kw}[1]{{\ensuremath {\mathsf{#1}}}\xspace}

\newcommand{\SpGAT} {\kw{SpGAT}}
\newcommand{\GAT} {\kw{GAT}}
\newcommand{\Cheby}{\kw{Cheby}}
\newcommand{\SpGATCheby}{\SpGAT-\Cheby}
\newcommand{\MAX}{\kw{MAX}}
\newcommand{\MEAN}{\kw{MEAN}}
\newcommand{\OctConv}{\kw{OctConv}}
\newcommand{\DeepWalk}{\kw{DeepWalk}}
\newcommand{\ChebyNet}{\kw{ChebyNet}}
\newcommand{\VanillaGCN}{\kw{Vanilla GCN}}
\newcommand{\GWNN}{\kw{GWNN}}
\newcommand{\ARMA}{\kw{ARMA}}
\newcommand{\GGNN}{\kw{GGNN}}
\newcommand{\GraphSAGE}{\kw{GraphSAGE}}

\newcommand{\HyperGraph}{\kw{HyperGraph}}
\newcommand{\HighOrder}{\kw{HighOrder}}
\newcommand{\AGG}{\kw{AGG}}
\newcommand{\SUM}{\kw{SUM}}
\newcommand{\Planetoid}{\kw{Planetoid}}

\newcommand{\GZoom}{\kw{GZoom(DGI)}}
\newcommand{\APPNP}{\kw{APPNP}}

\newcommand{\SGC}{\kw{SGC}}

\usepackage{float}

\AtBeginDocument{%
  \providecommand\BibTeX{{%
    \normalfont B\kern-0.5em{\scshape i\kern-0.25em b}\kern-0.8em\TeX}}}

\setcopyright{acmcopyright}
\copyrightyear{2021}
\acmYear{2021}
\acmDOI{10.1145/1122445.1122456}

\acmConference[DLG-KDD '21]{DLG-KDD '21: The Sixth International Workshop on Deep Learning on Graphs: Methods and Applications}{August 14-18, 2021}{Virtual Conference}
\acmBooktitle{DLG-KDD '21: The Sixth International Workshop on Deep Learning on Graphs: Methods and Applications,
  August 14-18, 2021, Virtual Conference}
\acmPrice{15.00}
\acmISBN{978-1-4503-XXXX-X/18/06}

\begin{document}

\title{Spectral Graph Attention Network with Fast Eigen-approximation}


\author{Heng Chang}
\authornote{Heng Chang is supported by 2020 Tencent Rhino-Bird Elite Training Program.}
\affiliation{%
  \institution{TBSI, Tsinghua University}
  }
\email{changh17@mails.tsinghua.edu.cn}

\author{Yu Rong}
\affiliation{%
  \institution{Tencent AI Lab}
}
\email{yu.rong@hotmail.com}

\author{Tingyang Xu}
\affiliation{%
  \institution{Tencent AI Lab}
}
\email{Tingyangxu@tencent.com}

\author{Wenbing Huang}
\affiliation{%
  \institution{Tsinghua University}
  }
\email{hwenbing@126.com}

\author{Somayeh Sojoudi}
\affiliation{%
  \institution{University of California at Berkeley}
  }
\email{sojoudi@berkeley.edu}

\author{Junzhou Huang}
\affiliation{%
  \institution{University of Texas at Arlington}
  }
\email{jzhuang@uta.edu}

\author{Wenwu Zhu}
\affiliation{%
  \institution{Tsinghua University}
  }
\email{wwzhu@tsinghua.edu.cn}

\renewcommand{\shortauthors}{Chang, Heng, et al.}

\begin{abstract}
  Variants of Graph Neural Networks (GNNs) for representation learning have been proposed recently and achieved fruitful results in various fields. Among them, Graph Attention Network (\GAT) first employs a self-attention strategy to learn attention weights for each edge in the spatial domain. However, learning the attentions over edges can only focus on the local information of graphs and greatly increases the computational costs. In this paper, we first introduce the attention mechanism in the spectral domain of graphs and present \textbf{Spectral Graph Attention Network (\SpGAT)} that learns representations for different frequency components regarding weighted filters and graph wavelets bases. In this way, \SpGAT can better capture global patterns of graphs in an efficient manner with much fewer learned parameters than that of \GAT. Further, to reduce the computational cost of \SpGAT brought by the eigen-decomposition, we propose a fast approximation variant \SpGATCheby. We thoroughly evaluate the performance of \SpGAT and \SpGATCheby in semi-supervised node classification tasks and verify the effectiveness of the learned attentions in spectral domain.
\end{abstract}



\maketitle

\section{Introduction}\label{sec:intro}

\begin{figure}[!t]
\centering
\subfigure[Original image]{
\includegraphics[width=0.31\columnwidth]{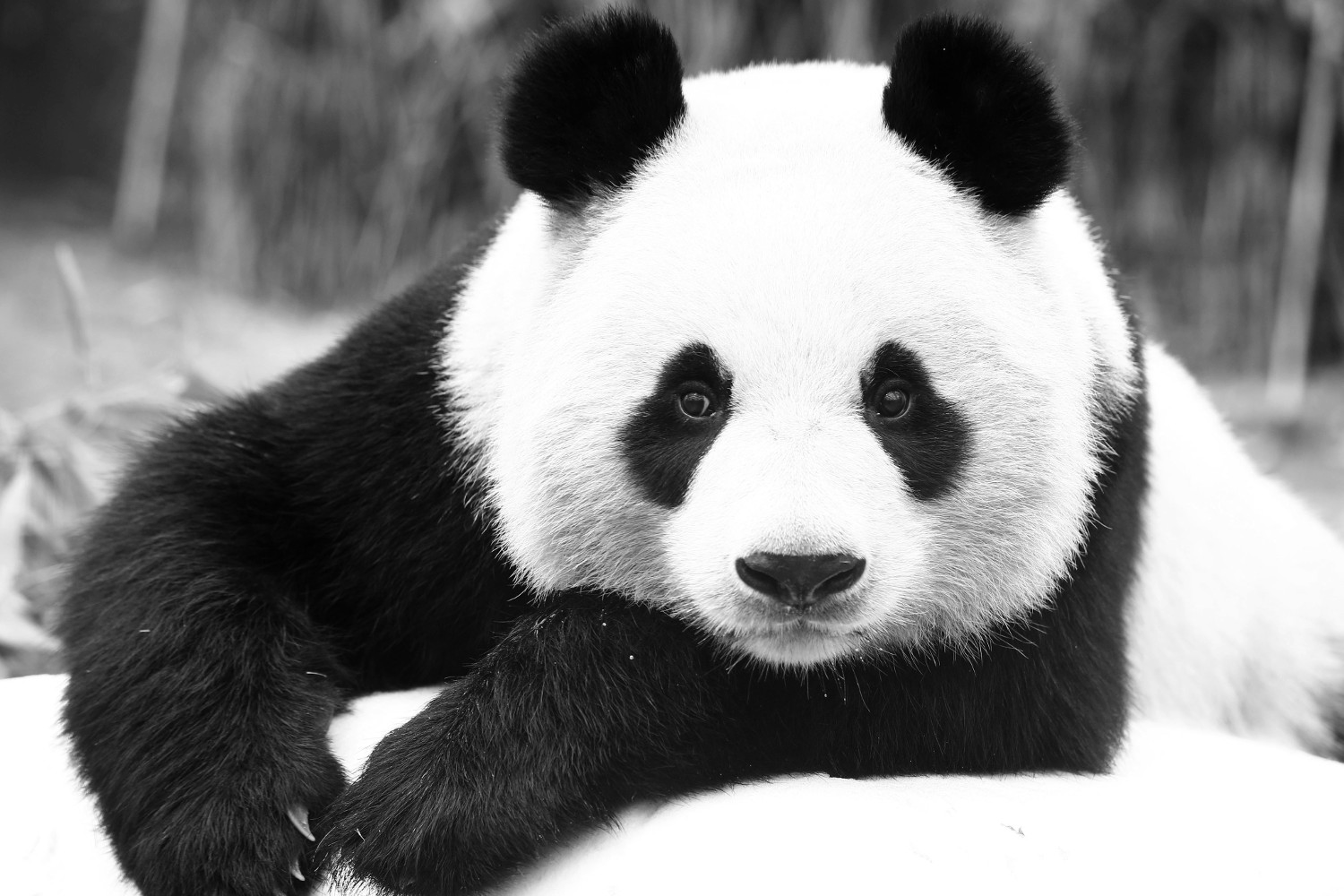}}
\subfigure[Low-frequency (\eg, background)]{
\includegraphics[width=0.31\columnwidth]{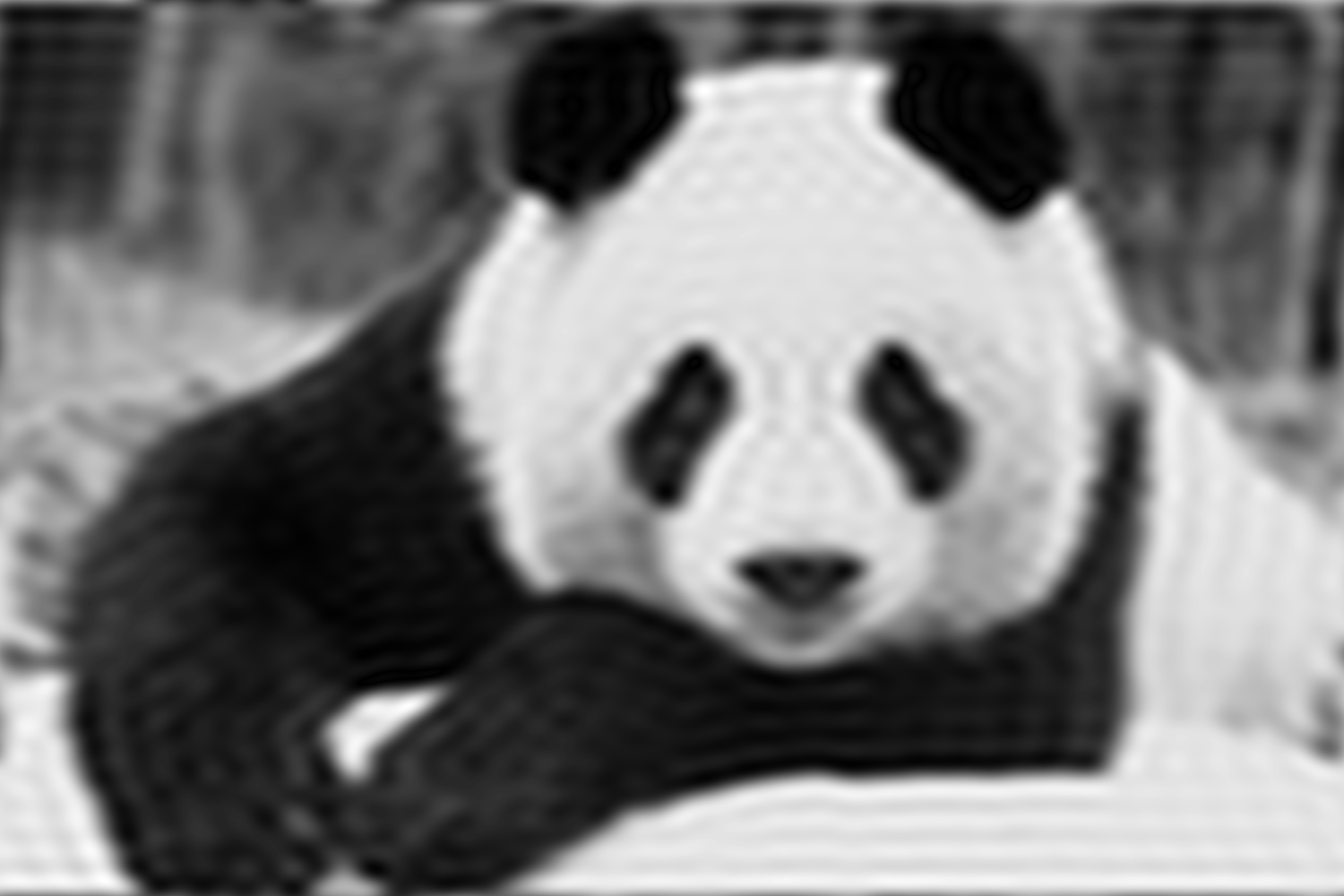}}
\subfigure[High-frequency (\eg, outlines)]{
\includegraphics[width=0.31\columnwidth]{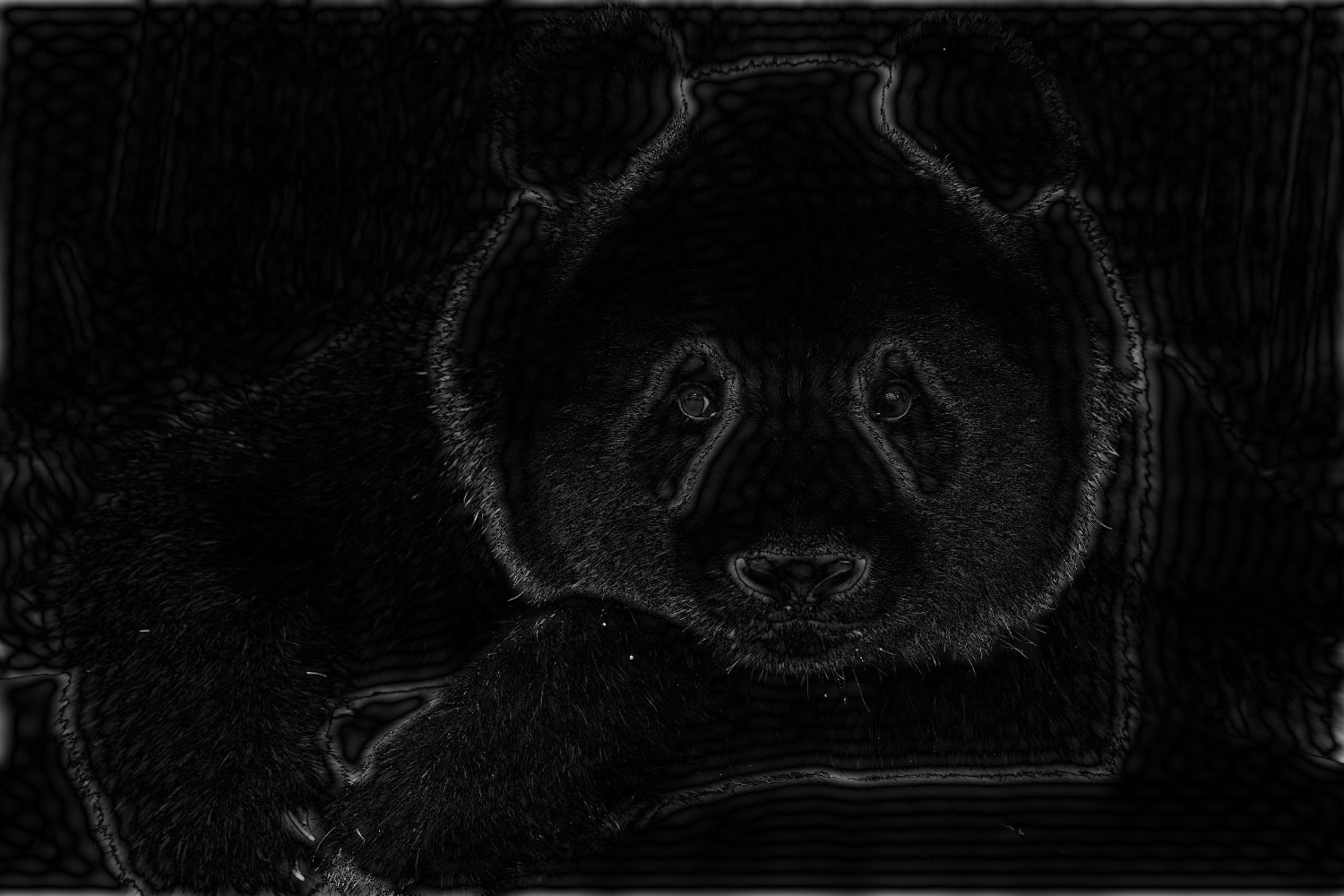}}
\subfigure[Original graph]{
\includegraphics[width=0.31\columnwidth]{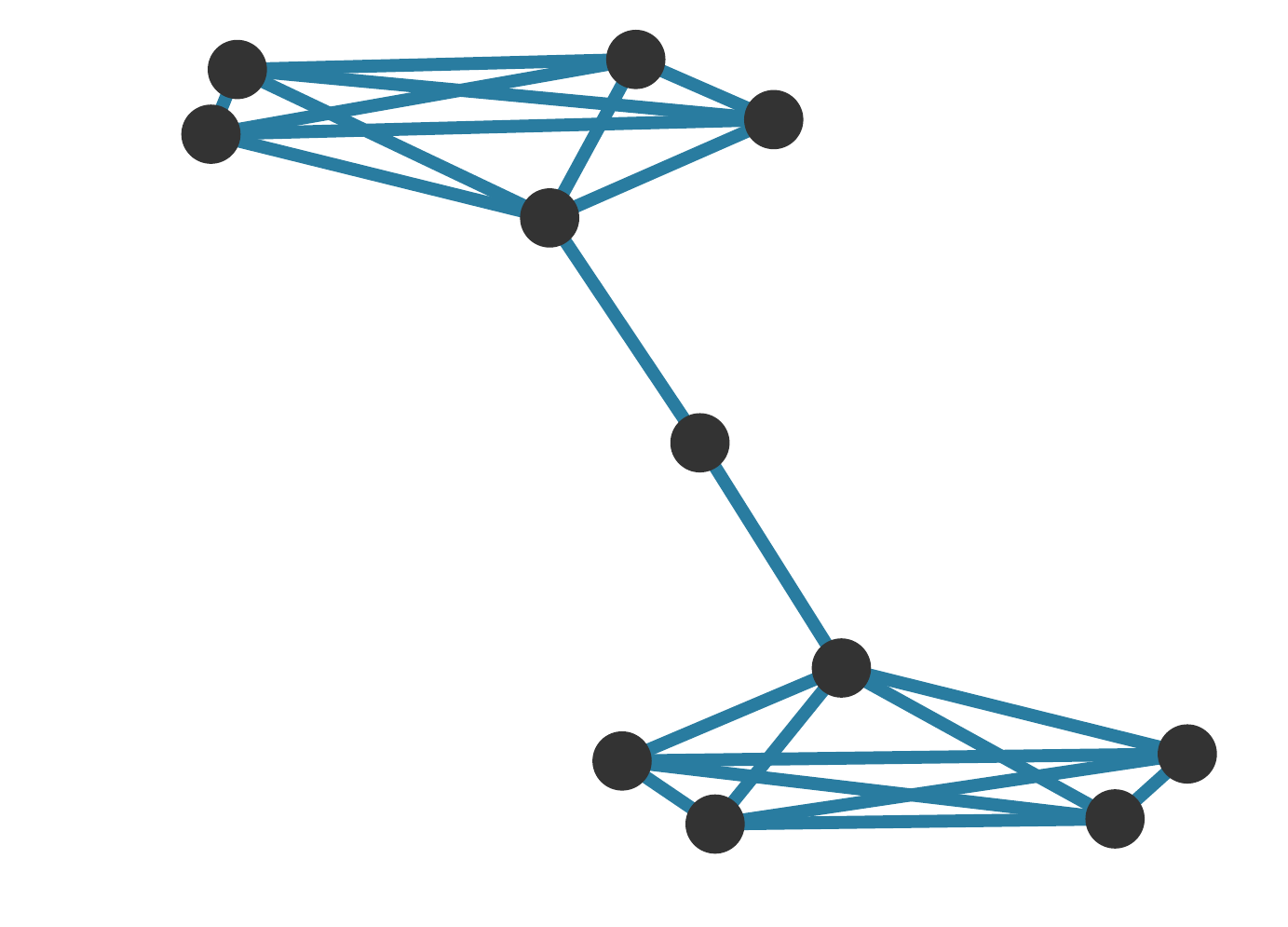}}
\subfigure[Low-frequency]{
\includegraphics[width=0.31\columnwidth]{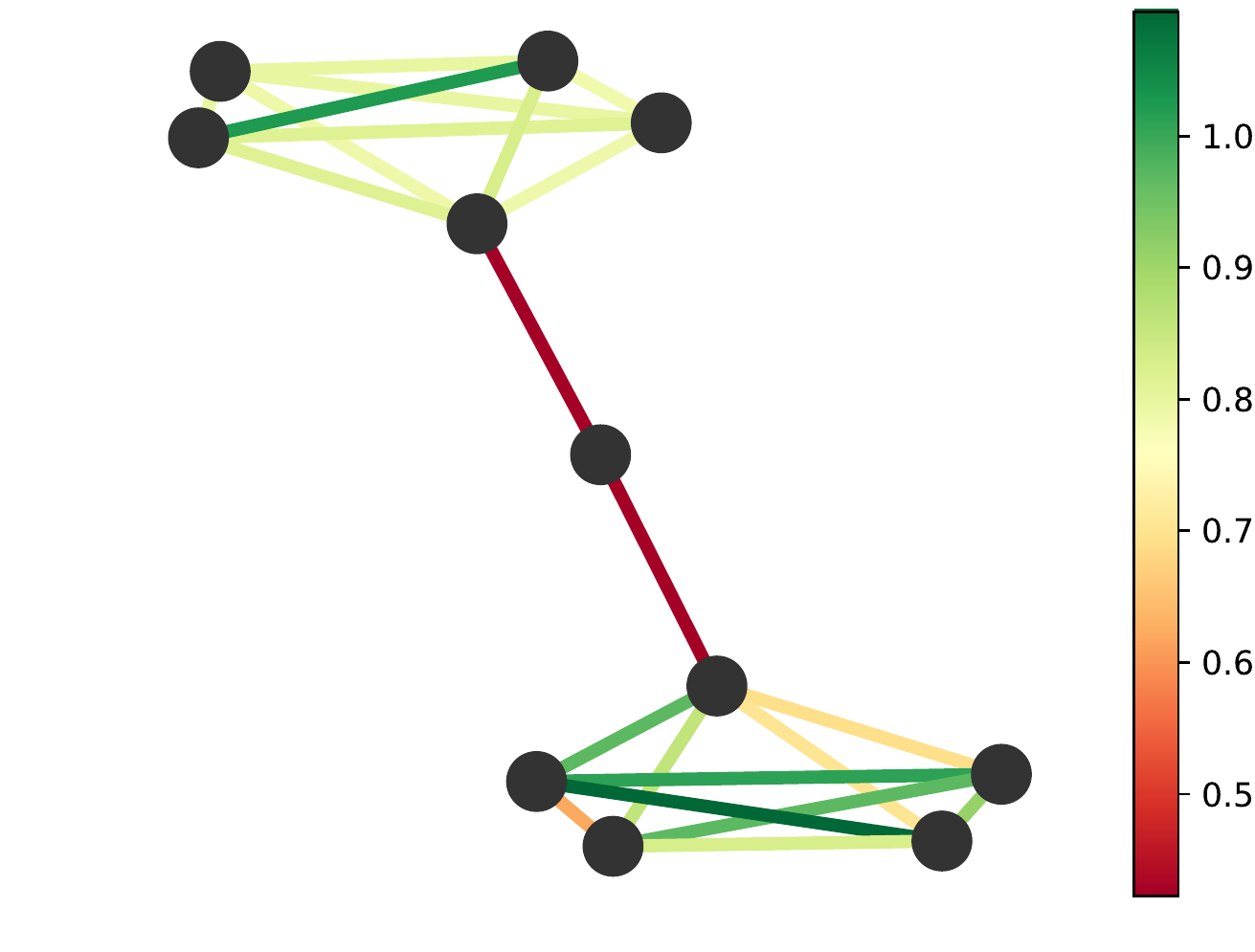}}
\subfigure[High-frequency]{
\includegraphics[width=0.31\columnwidth]{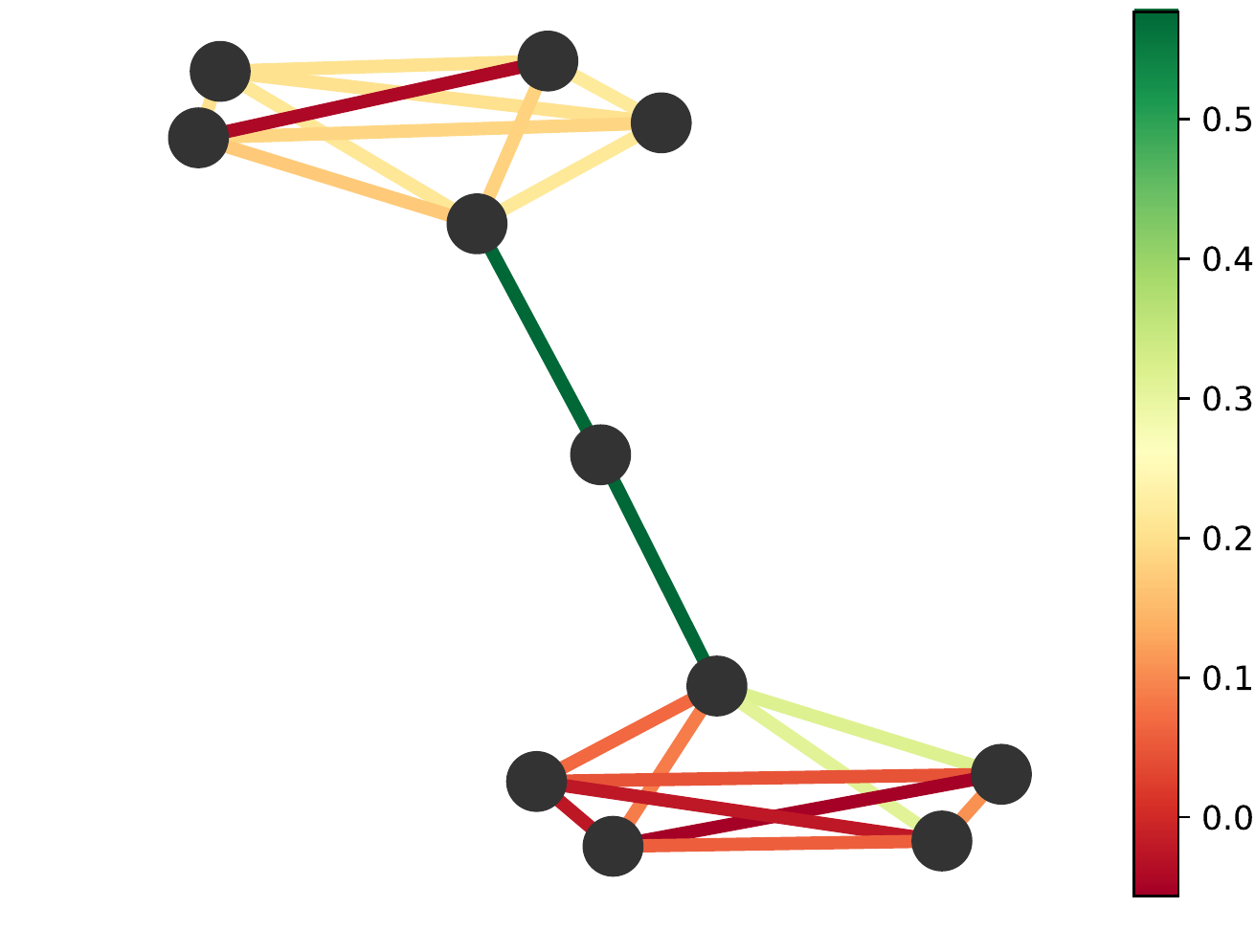}}
\vspace{-3mm}
\caption{Motivation: Separating the low- and high-frequency signals in both image and graph contributes to the feature learning.
Color bars in graphs indicate the measurement of reconstructed edge weights.
}
    \label{fig:cv and graph comparison.}
\vspace{-5mm}
\end{figure}

Graph Neural Networks (GNNs)~\cite{wu2019comprehensive} aim at imitating the expressive capability of deep neural networks from grid-like data (\eg, images and sequences) to graph structures. The fruitful progress of GNNs in the past decade has made them a crucial kind of tools for a variety of applications, from social networks~\cite{gu2020implicit}, computer vision~\cite{zeng2019graph}, to chemistry~\cite{liao2019lanczosnet}.

Graph Attention Network (\GAT)~\cite{velickovic2018gat}, as one central type of GNNs introduces the attention mechanism to further refine the convolution process in generic GCNs~\cite{ICLR2017SemiGCN}.
\GAT, along with its variants~\cite{gao2019graph, zhang2019adaptive, wang2019heterogeneous, wang2020direct}, considers the attention in a straightforward way: learning the edge attentions in the spatial domain. In this sense, this attention can capture the local structure of graphs, \ie, the information from neighbors. However, it is unable to explicitly encode the global structure of graphs. Furthermore, computing the attention weight for every edge in graphs is inefficient, especially for large graphs. 

In computer vision, a natural image can be decomposed into a low spatial frequency component containing the smoothly changing structure, \eg, background, and a high spatial frequency component describing the rapidly changing fine details, \eg, outlines~\cite{chen2019drop}. Figure~\ref{fig:cv and graph comparison.}(a) \textasciitilde ~\ref{fig:cv and graph comparison.}(c) depict the example of low- and high-frequency components on a panda image. Obviously, the contribution of different frequencies varies with respect to different downstream tasks. 

Similar pattern can be observed more naturally in graphs. According to graph signal processing (GSP), we can directly divide the low- and high-frequency components based on the ascending ordered eigenvalues of Laplacian in graphs. The eigenvectors associated with small eigenvalues carry smoothly varying signals, encouraging neighbor nodes to share similar values (local information). In contrast, the eigenvectors associated with large eigenvalues carry sharply varying signals across edges (global information)~\cite{donnat2018learning,maehara2019revisiting}. As demonstrated in Figure~\ref{fig:cv and graph comparison.}(d) \textasciitilde ~\ref{fig:cv and graph comparison.}(f), a barbell graph tends to retain the information inside the clusters when it is reconstructed with only low-frequency components(\ref{fig:cv and graph comparison.}(e)), but reserve knowledge between the clusters when constructed with only high-frequency ones (\ref{fig:cv and graph comparison.}(f)). 
Moreover, recent works~\cite{jin2019power,chang2020restricted} also reveal the different contributions of low- and high-frequency components in graphs to the learning of modern GNNs. 

\begin{figure}[tb]
\centering
\includegraphics [width=0.99\columnwidth]{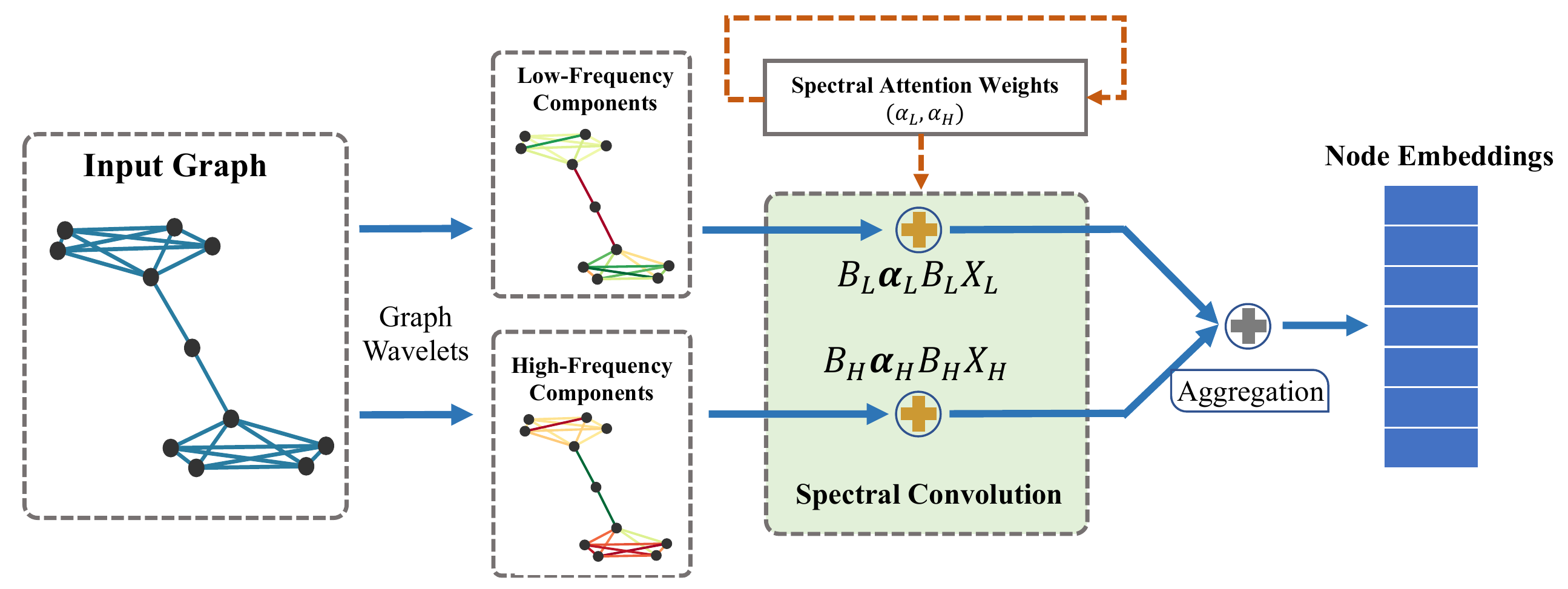}
\caption{Overview of proposed Spectral Graph Attention Network(\SpGAT). Two distinct convolutional kernels according to the low- and high-frequency components are constructed and the attention mechanism is employed on both kernels to capture the importance, respectively.}
 \vspace{-6mm}
\label{fig.octaveoverview}
\end{figure}

In this paper, to model the importance of low- and high-frequency components in graphs, we propose to extend the attention mechanism to the spectral domain. In this way, we can explicitly encode the structural information of graphs from a global perspective. Accordingly, we present Spectral Graph Attention Network (\SpGAT). In \SpGAT, we choose the graph wavelets as the spectral bases and decompose them into low- and high-frequency components with respect to their indices. Then we construct two distinct convolutional kernels according to the low- and high-frequency components and apply the attention mechanism on both kernels to capture their importance respectively. Finally, an pooling function as well as an activation function are applied to produce the output. Figure~\ref{fig.octaveoverview} provides an overview of the design of \SpGAT. Furthermore, we employ the Chebyshev polynomial approximation to compute the spectral wavelets of graphs and propose an variant \SpGATCheby, which is more efficient on large graphs. We thoroughly validate the performance of \SpGAT and \SpGATCheby on five benchmarks with 
fourteen competitive baselines. \SpGAT and \SpGATCheby achieve state-of-the-art results on all of the datasets.

\section{The Proposed Framework}
We denote $\gG=(\mathcal{V},\mathcal{E})$ as an undirected graph, where $|\mathcal{V}|=n$ is the set of $n$ nodes, and $\mathcal{E}$ is the set of edges, where $(v_i, v_j) \in \mathcal{E}$. The adjacency matrix is defined as $\bm{A} \in \mathbb{R}^{n\times n}$, where $\bm{A}_{ij} = 1$ indicates an edge $(v_i, v_j)$. We denote $\bm{D}$ as the degree matrix. $\hat{\bm{A}}=\hat{\bm{D}}^{-1/2}(\bm{A}+\bm{I})\hat{\bm{D}}^{-1/2}$ refers to the normalized adjacency matrix with self-loop, where $\bm{I}$ is the identity matrix and $\hat{\bm{D}} = \bm{D} + \bm{I}$.

From the spatial perspective, GNNs can usually be viewed as the feature aggregation among the neighbors of nodes in the spatial domain of graphs. Therefore, we write the feed-forward layer of GNNs in a general form:
\begin{align}
    \small
    \bm{h}_i' = \sigma {\big(} \AGG_{j\in \mathcal{N}_i}(\alpha_{ij}\bm{\Theta}\bm{h}_j) {\big)},
    \label{eq:gcn_general}
\end{align}
where $\mathcal{N}_i$ refers to the neighborhood set of node $i$ in graph. $\bm{H}'=\{\bm{h}'^{\text{T}}_1, \cdots, \bm{h}'^{\text{T}}_{n}\}$ are the output of hidden vectors from the layer with $\bm{H} = \{\bm{h}^{\text{T}}_1, \cdots, \bm{h}^{\text{T}}_{n}\}$ as the input features. $\sigma(\cdot)$ refers to the activation function, such as ReLU. $\bm{\Theta}^{\text{T}}\in \mathbb{R}^{p\times q}$ refers to the learning parameters of the layer, where $p$ and $q$ refer to the feature dimensions of input and output, respectively. 
$\alpha_{ij}$ refers to the aggregation weight of neighbor $j$ for node $i$. $\AGG(\cdot)$ refers to the aggregation function that aggregates the output of each neighbor, such as \SUM and \MEAN. For examples, \VanillaGCN can be viewed as the special case of Eq.~(\ref{eq:gcn_general}) where $\alpha_{ij}=\hat{\bm{A}}_{ij}$, and $\AGG(\cdot)=\SUM(\cdot)$. Meanwhile, \GAT proposes to compute the weight $\alpha_{ij}$ by a self-attention strategy and uses \SUM as aggregation function.

Other than the neighbor aggregation in the spatial domain~\cite{ICLR2017SemiGCN}, \VanillaGCN can also be understood from the perspective of GSP in the spectral domain:
\begin{equation}
  \small
  g_{\theta} \star \vx =  \bm{B} g_{\theta} \bm{B}^{\text{T}} \vx \, ,
\label{equ:fourier-conv}
\end{equation}
where $\vx$ is a signal on every node. $\bm{B}=\{\bm{b}_1,\cdots,\bm{b}_n\}$ are the spectral bases extracted from the graph. $g_{\theta}=\text{diag}(\theta)$ is a diagonal filter parameterized by $\theta$. Given Eq.~(\ref{equ:fourier-conv}), \VanillaGCN can be viewed as the spectral graph convolution based on the Fourier transformation on graphs with the first-order Chebyshev polynomial approximations \cite{ICLR2017SemiGCN}. Further, we can separate the spectral graph convolution into two stages \cite{ICLR2019GWNN}:
\begin{align}
    \small
    \notag\kw{feature}\text{ }\kw{transformation:}  & \bm{X} = \bm{H}\bm{\Theta^{\text{T}}},\\
    \kw{graph}\text{ }\kw{convolution:}  & \bm{H}' = \sigma(\bm{B}\bm{F}\bm{B}^{\text{T}}\bm{X}).\label{eq:twostage}
\end{align}

In Eq.~(\ref{eq:twostage}), $\bm{F}$ is a diagonal matrix for the kernel of graph convolution. For instance, the convolutional kernel for \VanillaGCN is $\bm{F}=\text{diag}(\lambda_1, \cdots, \lambda_n)$, where $\{\lambda_i\}_{i=1}^n$ are the eigenvalues of the normalized Laplacian $\bm{L}=\bm{I} - \hat{\bm{A}}$ in ascending order, while the spectral bases $\bm{B}$ for \VanillaGCN are the corresponding eigenvectors. 

\subsection{The Construction of \SpGAT Layer}
In this section, we start to describe the construction of \SpGAT layer. From the perspective of GSP, the diagonal values $(f_1, \cdots, f_n)$ on $\bm{F}$ can be treated as the \textbf{frequencies} on graphs when they equal to the eigenvalues. We denote the diagonal values with small / large indices as the low / high frequencies, respectively. Meanwhile, the corresponding spectral bases in $\bm{B}$ are low- and high-frequency components. As discussed in Section~\ref{sec:intro}, the low- and high-frequency components carry different structural information in graphs. In this vein, we first propose to split the spectral bases into two groups and re-write Eq.~(\ref{eq:twostage}) as follows:
\begin{align}
    \small
    \notag\bm{X_L} &= \bm{H}\bm{\Theta}_L^{\text{T}}, \bm{X}_H = \bm{H}\bm{\Theta}_H^{\text{T}},\\
    \bm{H}' &= \sigma \left(\AGG(\bm{B}_L\bm{F}_L\bm{B}_L^{\text{T}}\bm{X}_{L}, \bm{B}_H\bm{F}_H\bm{B}_H^{\text{T}}\bm{X}_{H}) \right),
    \label{eq:splitedhl}
\end{align}
where $\bm{B}_L=(\bm{b}_1,\cdots,\bm{b}_d)$ and $\bm{B}_H=(\bm{b}_{d+1},\cdots,\bm{b}_n)$ are the low- and high-frequency components, respectively. Here $d$ is a hyper-parameter that determines the splitting boundary of low- and high-frequency. When $\AGG(\cdot)=\SUM(\cdot)$, Eq.~(\ref{eq:splitedhl}) is equivalent to the graph convolution stage in Eq.~(\ref{eq:twostage}).

In Eq.~(\ref{eq:splitedhl}), $\bm{F}_L$ and $\bm{F}_H$ can also be viewed as the importances of the low- and high-frequency. Therefore, we introduce the learnable attention weights by exploiting the \emph{re-parameterization} trick:
\begin{align}
    \small
    \bm{H}' = \sigma \left(\AGG(\bm{B}_L\bm{\alpha}_L\bm{B}_L^{\text{T}}\bm{X}_{L}, \bm{B}_H\bm{\alpha}_H\bm{B}_H^{\text{T}}\bm{X}_{H}) \right).   
    \label{eq:splitedwithalpha}
\end{align}
In Eq.~(\ref{eq:splitedwithalpha}), $\bm{\alpha}_L=\text{diag}(\alpha_L,\cdots,\alpha_L)$ and $\bm{\alpha}_H=\text{diag}(\alpha_H,\cdots,\alpha_H)$ are parameterized by two learnable weights $\alpha_L$ and $\alpha_H$, respectively. To ensure $\alpha_L$ and $\alpha_H$ are positive and comparable, we normalize them by the $\mathrm{softmax}$ function in an attention manner:
\begin{align}
    \small
    \alpha_{\ast} = \mathrm{softmax}(\alpha_{\ast})=\frac{\exp{(\alpha_{\ast})} }{ \sum_{\ast} \exp(\alpha_{\ast} )}, \,\,\,\, \ast = L,H.\notag 
\end{align}
Theoretically, there are many approaches to re-parameterize $\bm{\alpha}_L$ and $\bm{\alpha}_H$, such as self-attention w.r.t the spectral basis $\bm{b}_i$. However, these kinds of re-parameterization can not reflect the nature of low- and high-frequency components. On the other hand, they may introduce too many additional learnable parameters, especially for large graphs. These parameters might prohibit the efficient training due to the limited amount of training data in graph learning, especially under the graph-based semi-supervised setting.

\subsection{Choice of Spectral Bases}
Another important design is the choice of the spectral basis. Instead of Fourier bases, we choose graph wavelets as spectral bases in \SpGAT following the observation on the advantages of spectral wavelets in recent works~\cite{donnat2018learning,ICLR2019GWNN}.
Formally, the wavelet on a graph $\psi_{si}(\lambda)$ is defined as the signal resulting from the modulation in the spectral domain of a signal $\vx$ centered around the associated node $i$~\cite{Hammond2011Wavelets,shuman2013GSP}. Then, given a graph $G$, the graph wavelet transformation is conducted by employing a set of wavelets $\bm{\Psi}_{s} = (\psi_{s1}(\lambda_{1}),\psi_{s2}(\lambda_{2}),\dots,\psi_{sn}(\lambda_{n}))$ as bases:
\begin{equation}
\small
\bm{\Psi}_s(\lambda) = \bm{U} g_{s}(\lambda) \bm{U}^\text{T},
\label{equ:wavelet}
\end{equation}
where $\bm{U}$ is the eigenvectors of the normalized Laplacian $\bm{L} = \bm{I} - \hat{\bm{A}}$. $g_s(\lambda) = \text{diag} \big(g_s(\lambda_1),g_s(\lambda_2),\dots,g_s(\lambda_{n})\big)$ is a scaling matrix with heat kernel scaled by hyperparameter $s$. The inverse of graph wavelets $\bm{\Psi}_s^{-1}(\lambda)$ is obtained by simply replacing $g_{s}(\lambda)$ with $g_{s}(-\lambda)$~\cite{donnat2018learning}. Smaller indices in graph wavelets correspond to low-frequency components and vice versa.
Overall, the architecture of \SpGAT layer with graph wavelet $\bm{\Psi}_s$ as bases can be written as:
\begin{align}
    \small
    \bm{X}  = \bm{H}\bm{\Theta^{\text{T}}}, \; \;
    \bm{H}' = \sigma \left( \AGG(\bm{\Psi}_{sL}\bm{\alpha}_L\bm{\Psi}_{sL}^{-1}\bm{X}, \bm{\Psi}_{sH}\bm{\alpha}_H\bm{\Psi}_{sH}^{-1}\bm{X}) \right).
    \label{eq:spgat}
\end{align}

\subsection{Parameter Complexity of \SpGAT}
In Eq.~(\ref{eq:spgat}), aiming to further reduce the parameter complexity, we share the parameters in feature transformation stage for $\bm{X}_L$ and $\bm{X}_H$, \ie, $\bm{\Theta}_L = \bm{\Theta}_H$. In this way, we reduce the parameter complexity from $\mathcal{O}(2 \times (p \times q + 1) )$ to $\mathcal{O}(p \times q + 2)$, which is nearly the same as \VanillaGCN, which is $\mathcal{O}(p\times q)$. The parameter complexity  of \SpGAT is much less than that of \GAT with $K$-head attention, which is $\mathcal{O}((p+2) \times q \times K)$. Comparing with \GAT, which captures the local structure of graphs from spatial domain, our proposed \SpGAT could better tackle global information by combining the low- and high-frequency features explicitly from spectral domain.

\section{Fast Approximation of \SpGAT}\label{sec.spgat.approx}

In \SpGAT, directly computing the transformation according to Eq.~(\ref{equ:wavelet}) is intensive for large graphs, since diagonalizing Laplacian $\bm{L}$ commonly requires $\mathcal{O}(n^{3})$ computational complexity. Fortunately, we can employ the Chebyshev polynomials to fast approximate the spectral graph wavelets without eigen-decomposition\cite{Hammond2011Wavelets}.  
\begin{theorem}
Let $s$ be the scaling parameter in the heat kernel $g_{s}(\lambda) = e^{-\lambda s}$, and $M$ be the degree of the Chebyshev polynomial approximation for the scaled wavelet (larger value of $M$ yields more accurate approximation but higher computational cost in the opposite), then the graph wavelet is given by
\begin{equation}
  \small
  \bm{\Psi}_{s}(\lambda) = \frac{1}{2} c_{0,s} + \sum_{i = 1}^{M} c_{i,s} T_{i}(\tilde{\bm{L}}), \;c_{i,s} = 2 e^{s} J_{i}(s),
\label{equ:wavelet_approximation}
\end{equation}
where $\tilde{\bm{L}} = \frac{2}{\lambda_{\text{max}}}\bm{L}-\bm{I}$, $T_{i}(\tilde{\bm{L}})$ is the $i_{th}$ order Chebyshev polynomial, and $J_{i}(s)$ is the modified Bessel function of the first kind.
\label{thm:fast approx}
\end{theorem}
Theorem~\ref{thm:fast approx} can be derived from Section~6 in~\cite{Hammond2011Wavelets}.
It should be noted that though \cite{ICLR2019GWNN} discusses the possibility to bring the method from~\cite{Hammond2011Wavelets} into approximating wavelets but with integral operations, we make the first 
attempt to integrate Theorem~\ref{thm:fast approx} into practice. Moreover, to accelerate the computation, we build a look-up table for the Bessel function $J_{i}(s)$ to avoid additional integral operations.

With this Chebyshev polynomial approximation, the computational cost of the spectral graph wavelet is decreased to $\mathcal{O}(M\| \mathcal{E}\| + Mn)$, where $\| \mathcal{E}\|$ is the total number of edges. Due to the real world graphs are usually sparse, this computational reduction can be very significant. We denote \SpGAT with Chebyshev polynomial approximation as \SpGATCheby.
As for \SpGATCheby, instead of using eigen-decomposition, we directly employ Eq.(\ref{equ:wavelet_approximation}) to speed up the computation of the spectral wavelets $\bm{\Psi}_{s}(\lambda)$. After that, the approximated $\bm{\Psi}_{s}(\lambda)$ are seamlessly fed into the original \SpGAT.

\section{Experiments}

\subsection{Experimental Setup}
Joining the practice of previous works, we mainly focus on five node classification benchmarks under semi-supervised setting with different graph size, feature type and public splitting, including three citation networks: Citeseer, Cora and Pubmed~\cite{Dataset2008Citeseer}, a coauthor network: Coauthor CS, and a co-purchase network: Amazon Photo~\cite{mcauley2015image}.
Statistical overview of all datasets can be found in~\cite{shchur2018pitfalls}.

We thoroughly evaluate the performance of \SpGAT with fourteen representative baselines:
\begin{itemize}
    \item \textbf{Traditional graph embedding methods}: \DeepWalk~\cite{perozzi2014deepwalk} and \Planetoid~\cite{yang2016revisiting};
    \item \textbf{Spectral-based GNNs}: \ChebyNet~\cite{Defferrard2016ChebNet},  \VanillaGCN~\cite{ICLR2017SemiGCN},\SGC \cite{wu2019simplifying}, \GWNN~\cite{ICLR2019GWNN}, \ARMA~\cite{ARMAbianchi2019graph}, and \GZoom~ \cite{deng2020graphzoom};
    \item \textbf{Spatial-based GNNs}: \GGNN~\cite{GGNNli2016gated}, \GraphSAGE~\cite{Hamilton2017Inductive}, \GAT~\cite{velickovic2018gat}, \HyperGraph~\cite{HyperGraphbai2019hypergraph}, \HighOrder~\cite{HighOrdermorris2019weisfeiler}, and \APPNP~\cite{klicpera_predict_2019}.
\end{itemize}

For all experiments, a 2-layer neural network is constructed using TensorFlow~\cite{abadi2015tensorflow} with 64 hidden units. We train our model utilizing the Adam optimizer~\cite{kingma2014adam} with an initial learning rate $lr = 0.01$. Early stopping is used with a window size of 100. 
Most training processes are stopped in less than 200 steps as expected. 
We initialize the weights following~\cite{glorot2010understanding}, employ $5 \times 10^{-4}$ L2 regularization and dropout the input and hidden layers to prevent overfitting~\cite{srivastava2014dropout}.
For constructing wavelets, we set $s=1$, $t=1\times 10^{-4}$ for \SpGAT, and $M = 1$, $s = 2$ and $t=1\times 10^{-4}$ for \SpGATCheby on all datasets.
In addition, we employ the grid search to determine the best $d$ of low-frequency components and the impact of this parameter would be discussed in Section~\ref{sec.abstudy}. 
Two variants with \MEAN-pooling and \MAX-pooling are implemented to demonstrate the effectiveness of aggregation function in \SpGAT and \SpGATCheby. Without other specification, we use \MAX-pooling in both models.

\subsection{Semi-supervised Node Classification}
Table~\ref{tab:benchmark_performance} summaries the results on all datasets. For all baselines, we reuse the results from their public literature. From Table~\ref{tab:benchmark_performance}, we have these findings: (1) Clearly, the attention-based GNNs (\GAT, \SpGAT and \SpGATCheby) achieve relatively better performance across all datasets. It validates that the attention mechanism can capture the important patterns from either spatial or spectral perspective.
(2) Specifically, \SpGAT and \SpGATCheby achieve the best performance across all datasets.
Particularly on Coauthor CS, the best accuracy by \SpGATCheby-\MAX is $92.5\%$ and it is better than the previous best ($90.7\%$), which is regarded as a remarkable boost considering the challenge on this benchmark. (3) Compared with \MEAN aggregation, \MAX aggregation seems to be a better choice for both models. This may due to that \MAX aggregation can preserve the significant signals learned by \SpGAT. (4) It is worthy to note that to achieve such results, both \SpGAT and \SpGATCheby only employ the attention on low- and high-frequency of graphs in spectral domain, while \GAT needs to learn the attention weights on every edge in spatial domain. It verifies that \SpGAT is more efficient than \GAT, since the global information of graphs can be better captured from spectral domain while with less parameters.

\begin{table}[t]

    \caption{Experimental results (in percentage) on semi-supervised node classification. }
    \vspace{-4mm}
    \small
    \centering
    \resizebox{0.49\textwidth}{!}{%
        \begin{tabular}[t]{lccccc}
            \toprule
            \textbf{Model}  & \textbf{Citeseer} & \textbf{Cora} & \textbf{Pubmed} & \textbf{Coauthor CS} & \textbf{Amazon Photo} \\
            \hline
            \DeepWalk \cite{perozzi2014deepwalk} & $43.2$ & $67.2$ & $65.3$ & $-$ & $-$ \\ 
            
            \Planetoid \cite{yang2016revisiting} & $64.7$ & $75.7$ & $77.2$ & $-$ & $-$ \\ 
            
            \hline

            \ChebyNet \cite{Defferrard2016ChebNet} & 
            $69.8$ &
            $81.2$ &
            $74.4$ &
            $90.5$ & $89.6$ \\ 
            
            \VanillaGCN \cite{ICLR2017SemiGCN} & 
            $70.3$ &
            $81.5$ &
            $79.0$ &
            $89.8$ & $90.6$ \\ 
            
            \GWNN \cite{ICLR2019GWNN} & 
            $71.7$ &
            $82.8$ &
            $79.1$ &
            $90.3$ & $88.5$ \\ 
            
            \ARMA \cite{ARMAbianchi2019graph} & 
            $70.9$ &
            $83.3$ &
            $78.4$ &
            $90.6$ & $86.4$ \\ 
            
            \SGC \cite{wu2019simplifying} & $71.9$ & $81.0$ & $78.9$ & $89.3$ & $90.1$ \\
            \GZoom \cite{deng2020graphzoom} & $71.7$ & $83.2$ & $77.1$ & $88.9$ & $89.3$ \\
            \hline
            
            \GGNN \cite{GGNNli2016gated} & 
            $64.6$ &
            $77.6$ &
            $75.8$ &
            $86.6$ & $74.1$ \\ 
            
            \GraphSAGE \cite{Hamilton2017Inductive} & 
            $67.2$ &
            $74.5$ &
            $76.8$ &
            $90.1$ & $90.1$ \\ 
            
            
            
            \GAT \cite{velickovic2018gat} & 
            $72.5$ &
            $83.0$ &
            $79.0$ &
            $85.5$ & $89.7$ \\ 
            
            \HyperGraph \cite{HyperGraphbai2019hypergraph} & 
            $71.2$ & 
            $82.7$ & 
            $78.4$ & $86.9$ & $87.5$ \\

            \HighOrder \cite{HighOrdermorris2019weisfeiler} &
            $64.2$ &
            $76.6$ &
            $75.0$ &
            $84.2$ & $26.1$ \\

            \APPNP \cite{klicpera_predict_2019} & $\textbf{72.7}$ & $83.1$ & $79.1$ & $90.7$ & $91.8$ \\

            \hline
            \hline
             
            \SpGAT-\MEAN &  $71.6 \pm 0.2$ & $82.6 \pm 0.3 $ & $ 80.3 \pm 0.2 $ & $ 91.0 \pm 0.3 $ & $ 91.8 \pm 0.3 $ \\ 
            \SpGAT-\MAX  &  $\textbf{72.1} \pm 0.2$ & $\textbf{83.7} \pm 0.2 $ & $ \textbf{80.6} \pm 0.3 $ & $ 91.6 \pm 0.3 $ & $ 91.4 \pm 0.2 $ \\ 
            \SpGATCheby-\MEAN &  $70.0 \pm 0.2$ & $80.7 \pm 0.4 $ & $ 78.3 \pm 0.3 $ & $ 91.1 \pm 0.2 $ & $ 92.4 \pm 0.1 $ \\ 
            \SpGATCheby-\MAX &  $71.1 \pm 0.4$ & $82.1 \pm 0.3 $ & $ 80.2 \pm 0.2 $ & $ \textbf{92.1} \pm 0.1 $ & $ \textbf{92.8} \pm 0.2 $ \\
            \bottomrule
        \end{tabular}
        }
\label{tab:benchmark_performance}
\vspace{-4mm}
\end{table}

\begin{figure}[t]
\centering
\subfigure {\includegraphics[width=0.32\columnwidth]{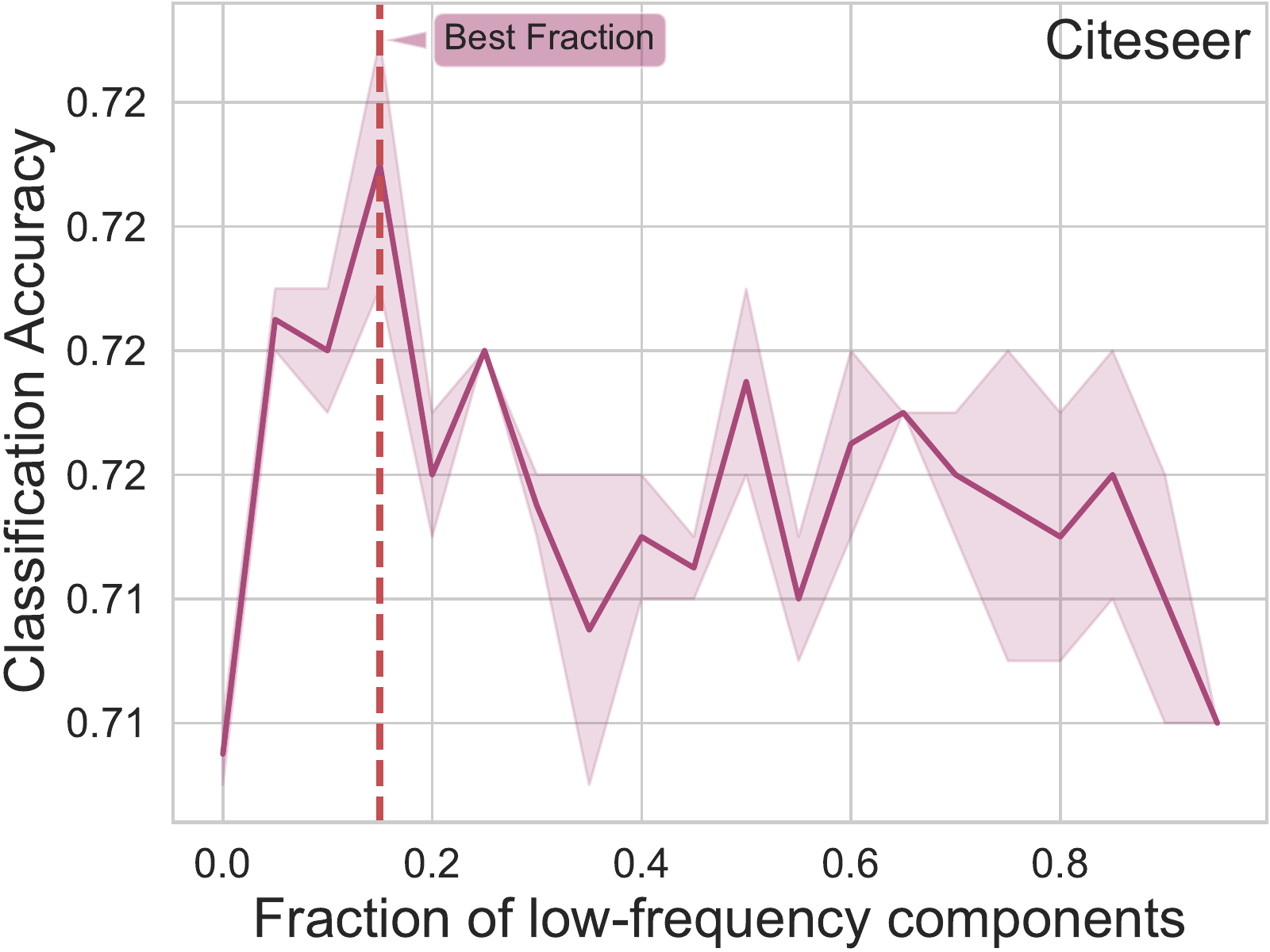}}
\subfigure {\includegraphics[width=0.32\columnwidth]{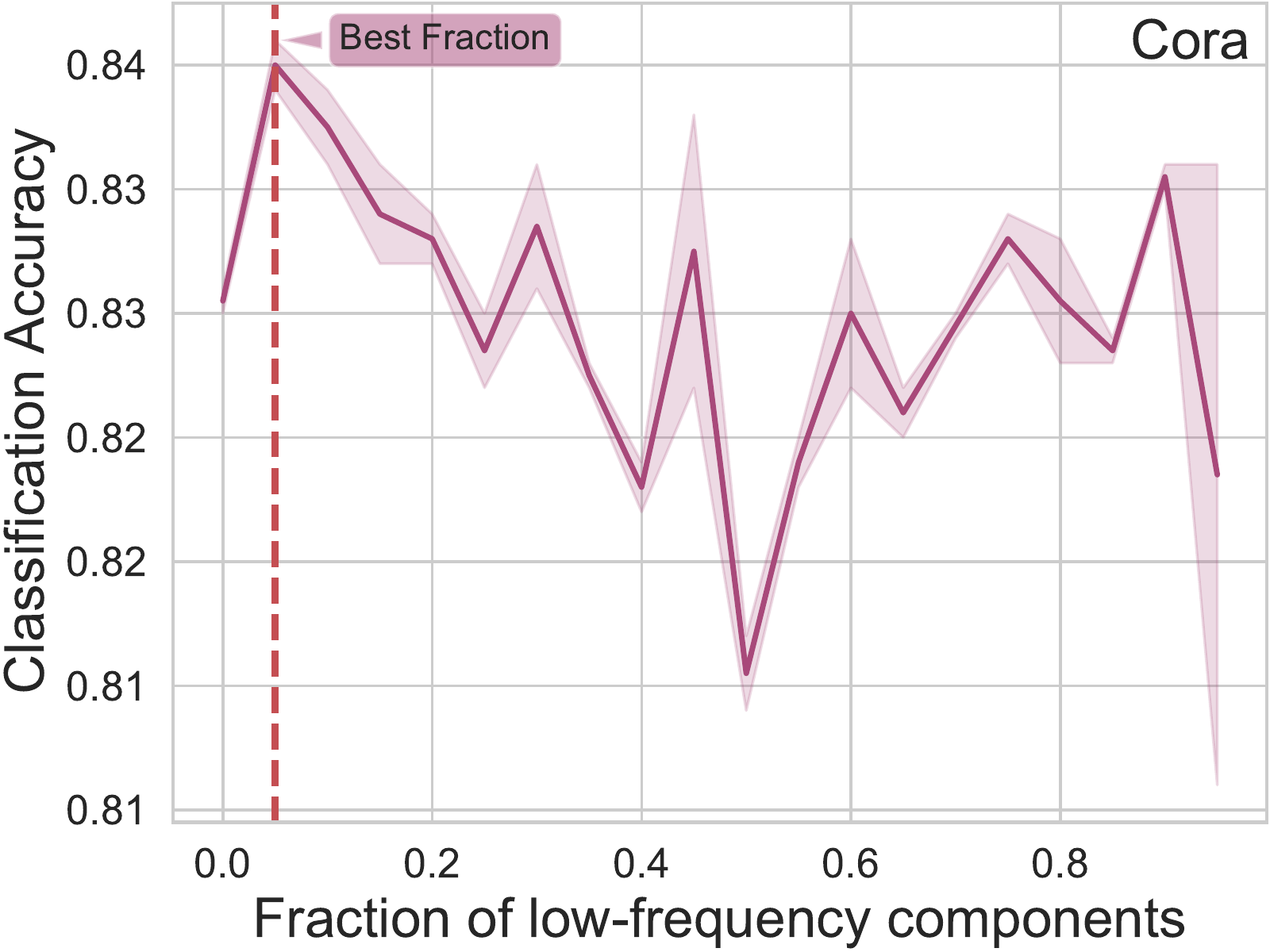}}
\subfigure {\includegraphics[width=0.32\columnwidth]{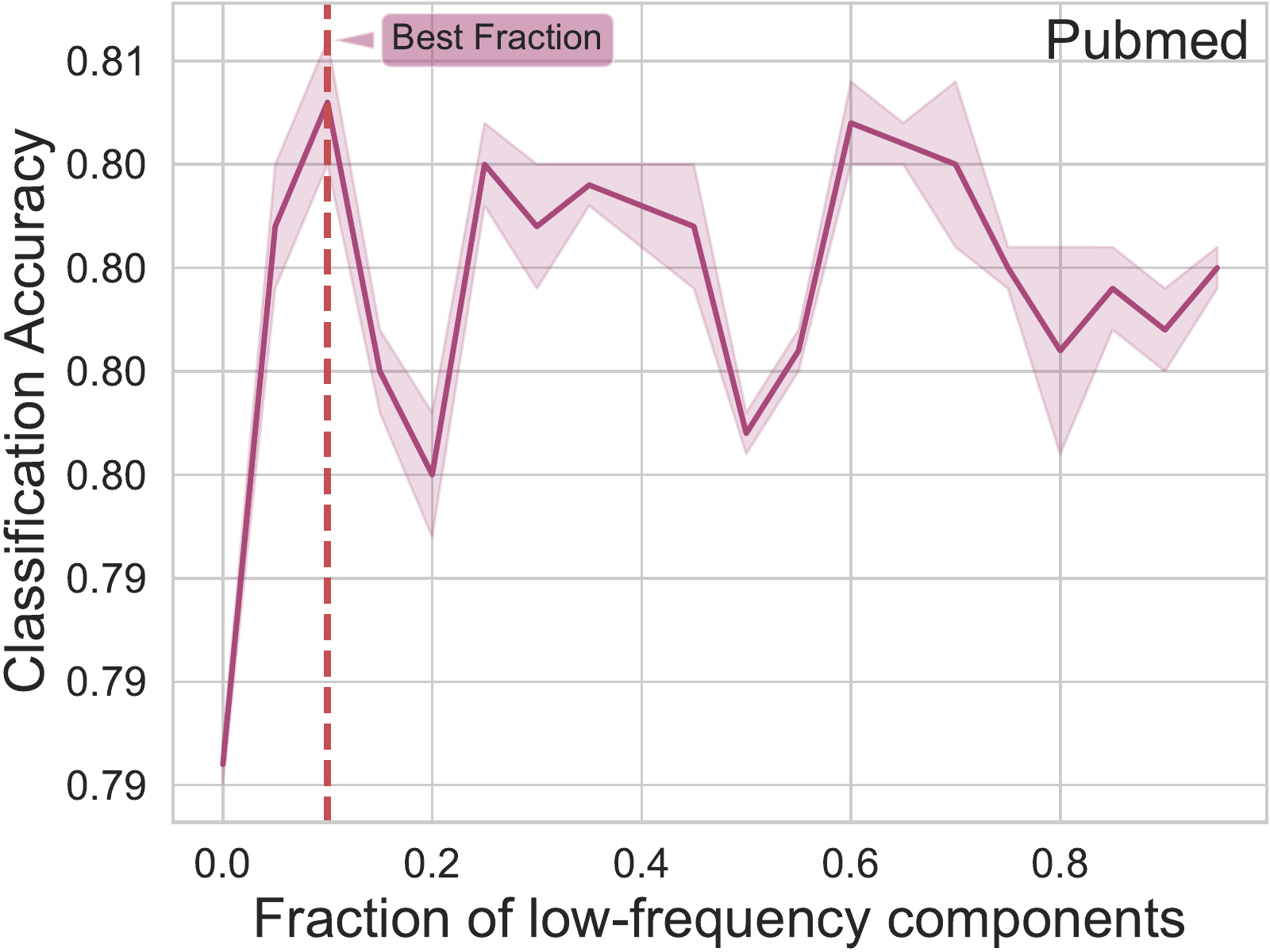}}
\vspace{-4mm}
\caption{The performance of \SpGAT w.r.t the proportion of low-frequency $d$ on citation datasets. The best fraction is marked with the dashed vertical line.}
    \label{fig:fraction}
\vspace{-5mm}
\end{figure}

\subsection{Ablation Studies}\label{sec.abstudy}

\subsubsection{The Impact of Proportion of Low-frequency Components  $d$} \label{sec.abstudy.impactdn}
To evaluate the impact of the hyperparameter $d$, we fix the other hyperparameters and vary $d$ from $0$ to $100\%$ linearly to run \SpGAT on citation datasets. Figure~\ref{fig:fraction} depicts the mean (in bold line) and variance (in light area) of every $d$.
As shown in Figure~\ref{fig:fraction}, the mean value curve of three datasets exhibits similar pattern, that is, the best performance is achieved when $d$ is small. The best proportions of low-frequency components are $15\%$, $5\%$ and $10\%$ for Citeseer, Cora and Pubmed, respectively. In the other words, 
consistently, only a relatively small fraction of components needs to be treated as low-frequency components in \SpGAT. This finding is consistent with the argument from~\cite{maehara2019revisiting}, which has discussed that a small fraction of low-frequency components already contains sufficient information for the reconstruction of signals. Thus it might be a good choice to select a small $d$ for our model when generalizing to other datasets. A theoretically heuristic method to determine $d$ could be interesting and will be left for future exploration.

\subsubsection{Time Efficiency of \SpGAT and \SpGATCheby}
As discussed in Section ~\ref{sec.spgat.approx}, 
we propose the fast approximation of spectral wavelets $\bm{\Psi}_{s}(\lambda)$ according to Chebyshev polynomials. 
To elaborate its efficiency, we compare the time cost of calculating $\bm{\Psi}_{s}(\lambda)$ between via eigen-decomposition (\SpGAT) and fast approximation (\SpGATCheby).
We report the mean time cost of \SpGAT and \SpGATCheby with second-order Chebyshev polynomials after 10 runs for citation datasets. As shown in Table~\ref{tab:timeCost}, we can find that this fast approximation can greatly accelerate the training process. Specifically, \SpGATCheby runs $ 7.9 \times$ times faster than \SpGAT for obtaining $\bm{\Psi}_{s}(\lambda)$ on the relatively large dataset Pubmed. It further validates the scalability of the fast approximation approach.

\begin{table}
    \renewcommand{\arraystretch}{1.11}
    \centering
    \small
    \caption{Running time ($s$) comparison for obtaining spectral wavelets $\bm{\Psi}_{s}(\lambda)$ between \SpGAT and \SpGATCheby. \label{tab:timeCost}}
    \vspace{-3mm}
    \resizebox{0.75\columnwidth}{!}{%
    \begin{tabular}{ c c c }
    \hline
    Models  & Eigen-decomposition & Fast approximation \\
    \hline
    Citeseer    & $11.23$      &  $5.19$ (\textasciitilde $ \textbf{2.2} \times$)   \\
    \hline
    Cora    & $5.79$      &  $2.78$ (\textasciitilde $ \textbf{2.1} \times$)   \\
    \hline
    Pubmed    & $1185.12$  &  $150.79$ (\textasciitilde $ \textbf{7.9} \times$)   \\
    \hline
    \end{tabular}
    }
    \vspace{-3mm}
\end{table}

\begin{figure}[htb]
\centering
\includegraphics [width=0.7\columnwidth]{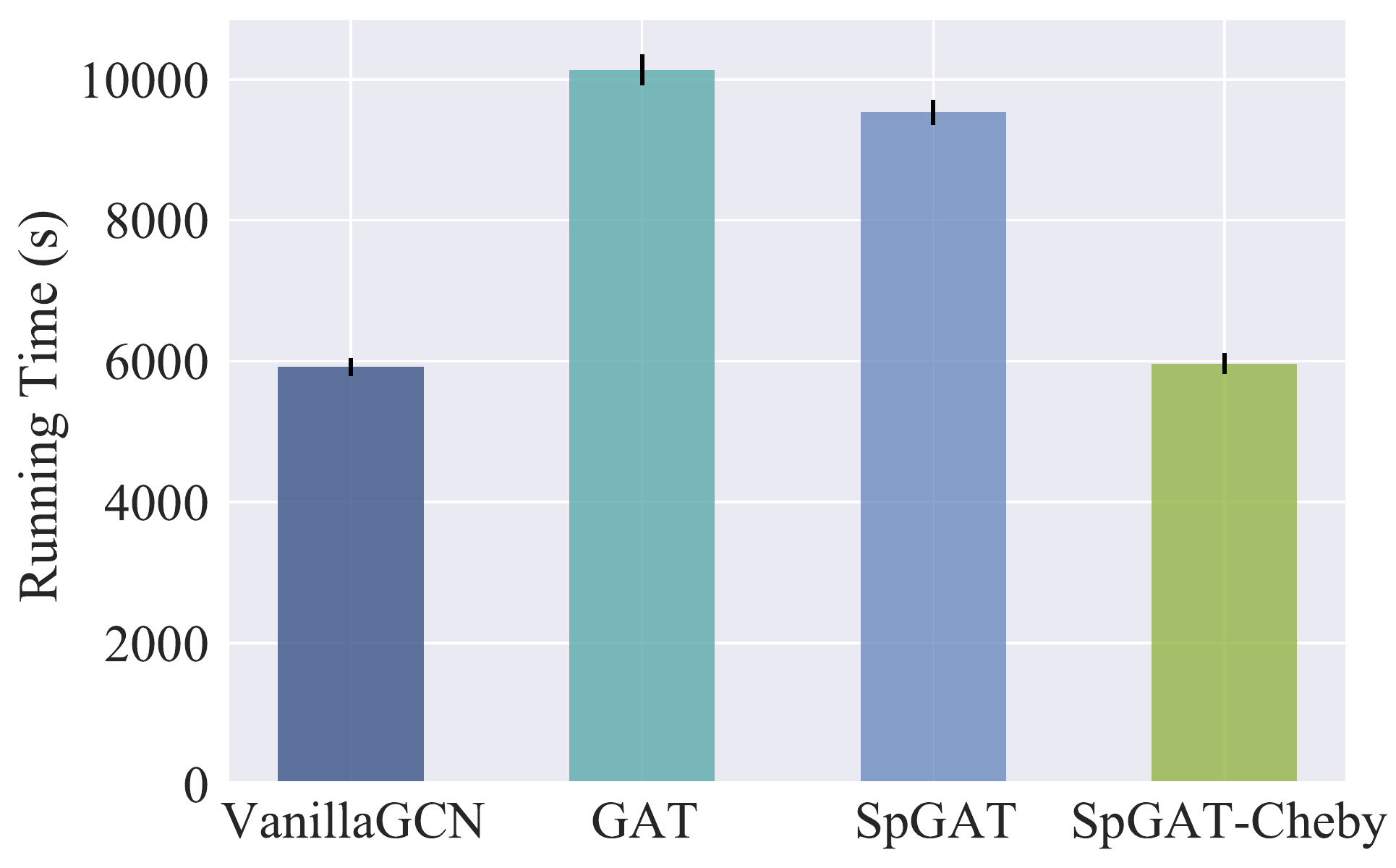}
\vspace{-4mm}
\caption{Running time ($s$) comparison on Pubmed.}
\vspace{-4mm}
\label{fig:running_time with GAT}
\end{figure}

Meanwhile, we also compare the running time cost of 200 epochs among our proposed methods, \VanillaGCN and \GAT on Pubmed dataset, and the results can be found in Figure~\ref{fig:running_time with GAT}. We can observe that \SpGAT runs slightly faster than \GAT. Furthermore, with the fast approximation technique, \SpGATCheby saves nearly half of the time comparing with \GAT and achieves comparable efficiency against \VanillaGCN. It further confirms the efficiency of the proposed approximation on spectral wavelets.

\section{Conclusion}
In this paper, we propose \SpGAT, a novel spectral-based graph convolutional neural network to learn the representation of graphs with respect to different frequency components in the spectral domain. 
By introducing the distinct trainable attention weights for low- and high-frequency components, \SpGAT can effectively capture both local and global information in graphs and enhance the performance of GNNs. Furthermore, a fast variant \SpGATCheby based on Chebyshev polynomial approximation is proposed to accelerate the spectral graph wavelets calculation and benefit the scalability. To the best of our knowledge, this is the first attempt to adopt the attention mechanism to the spectral domain of graphs. It is expected that \SpGAT and \SpGATCheby could shed light on building more efficient architectures for the area of graph learning.

\small{
\bibliographystyle{ACM-Reference-Format}
\bibliography{spgat}
}
\appendix

\section{Datasets}

Joining the practice of previous works, we mainly focus on five node classification benchmark datasets under semi-supervised setting with different graph size and feature type.
(1) Three citation networks: Citeseer, Cora and Pubmed~\cite{Dataset2008Citeseer}, which aims to classify the research topics of papers. (2) A coauthor network: Coauthor CS which aims to predict the most active fields of study for each author from the KDD Cup 2016 challenge\footnote{https://kddcup2016.azurewebsites.net}. (3) A co-purchase network: Amazon Photo~\cite{mcauley2015image} which aims to predict the category of products from Amazon.
For the citation networks, we follow the public split setting provided by~\cite{yang2016revisiting}, that is, 20 labeled nodes per class in each dataset for training and 500 / 1000 labeled samples for validation / test respectively. For the other two datasets, we follow the splitting setting from~\cite{shchur2018pitfalls,chen2019measuring}.
Statistical overview of all datasets is given in Table~\ref{tab:datasets}. Label rate denotes the ratio of labeled nodes fetched in training process.

\begin{table}[htp]
\centering
\caption{\label{tab:datasets}The overview of dataset statistics.}
\resizebox{\columnwidth}{!}{%
\begin{tabular}{l r r r r r}
\toprule
\textbf{Dataset} & \textbf{Nodes} & \textbf{Edges} & \textbf{Classes} & \textbf{Features} & \textbf{Label rate}  \\[0.05em]\hline \\[-0.8em]
\textbf{Citeseer} & 3,327 & 4,732 & $6$ & 3,703 & $0.036$ \\
\textbf{Cora} & 2,708 & 5,429 & $7$ & 1,433 & $0.052$ \\
\textbf{Pubmed} & 19,717 & 44,338 & $3$ & 500 & $0.003$ \\
\textbf{Coauthor CS} & 18,333 & 81,894 & $15$ & 6,805 & $0.016$ \\
\textbf{Amazon Photo} & 7,487 & 11,9043 & $8$ & 745 & $0.021$ \\
\bottomrule
\end{tabular}
}
\end{table}

\section{Full Ablation Studies}

\subsection{The Learned Attention on Low- and High-frequency Components}
In this Section, we also show how the learned attentions of \SpGAT w.r.t the best proportion for Citeseer, Cora and Pubmed which are demonstrated in Table~\ref{tab:learned alpha weights}. Interestingly, despite the small proportion, the attention weight of low-frequency components learned by \SpGAT is much larger than that of high-frequency components in each layer consistently. Hence, \SpGAT is successfully to capture the importance of low- and high-frequency components of graphs in the spectral domain. Moreover, as pointed out by~\cite{donnat2018learning,maehara2019revisiting}, the low-frequency components in graphs usually indicate smooth varying signals which can reflect the locality property in graphs. It implies that the local structural information is important for these datasets. This may explain why \GAT also gains good performance on these datasets.

\begin{table}
\caption{Learned attention weights $\alpha_{L}$ and $\alpha_{H}$ of  \SpGAT for low- and high-frequency w.r.t the best proportion of low frequency components $d$ (number followed after the name of datasets). \label{tab:learned alpha weights}}
\vspace{-2mm}
\centering
\resizebox{0.49\textwidth}{!}{%
\begin{tabular}{@{} lcc|cc|cc@{}}
\toprule
    Dataset & \multicolumn{2}{c|}{Citeseer ($15\%$)} & \multicolumn{2}{c|}{Cora ($5\%$)} & \multicolumn{2}{c}{Pubmed ($10\%$)}  \\
\midrule
Attention filter weights & $\alpha_{L}$  & $\alpha_{H}$   & $\alpha_{L}$  & $\alpha_{H}$  & $\alpha_{L}$  & $\alpha_{H}$   \\
\midrule
Learned value (first layer) & $\textbf{0.84}$ &  $0.16$ &   $\textbf{0.72}$ & $0.23$ & $\textbf{0.86}$ & $0.14$ \\
\midrule
Learned value (second layer) & $\textbf{0.94}$ &  $0.06$ &  $\textbf{0.93}$ & $0.07$ & $$\textbf{0.93}$$ & $0.07$ \\
\bottomrule
\end{tabular}
}
\end{table}

\begin{table}[htb]
\caption{The results of ablation study on low- and high-frequency components. \label{tab:test only with L/H}}
\vspace{-2mm}
\centering
\resizebox{\columnwidth}{!}{
\begin{tabular}{@{} lccc@{}}
\toprule
    Methods & {Citeseer ($15\%$)} & {Cora ($5\%$)} & {Pubmed ($10\%$)}  \\
\midrule
with low-frequency & $57.7$ & $66.8$ & $76.7$ \\
\midrule
with high-frequency & $70.9$ & $82.4$ & $80.4$ \\
\midrule
\SpGAT & $\textbf{72.3}$ & $\textbf{83.8}$ & $\textbf{80.8}$ \\
\bottomrule
\end{tabular}
}
\vspace{-1mm}
\end{table}

\subsection{Only Low- and High-frequency Components}
To further elaborate the importance of low- and high-frequency components in \SpGAT, we conduct the ablation study on the classification results by testing only with low- or high-frequency components w.r.t the best proportion. Specially, we manually set $\alpha_{L}$ or $\alpha_{H}$ to 0 during testing stage to observe how the learned low- and high-frequency components in graphs affect the classification accuracy. From Table~\ref{tab:test only with L/H}, we can observe that: 
\begin{itemize}
    \item  Both low- and high-frequency components are essential for the model. Since removing any components downgrade the over performance.
    \item \SpGAT with very small proportion (5\% - 15\%) of low-frequency components can achieve the comparable results to those obtained by full \SpGAT. It reads that the low-frequency components contain more information that can contribute to the feature representation learned from the model.
\end{itemize}

\begin{figure*}[!t]
\centering
\subfigure [\VanillaGCN on Citeseer] {\includegraphics[width=0.245\linewidth]{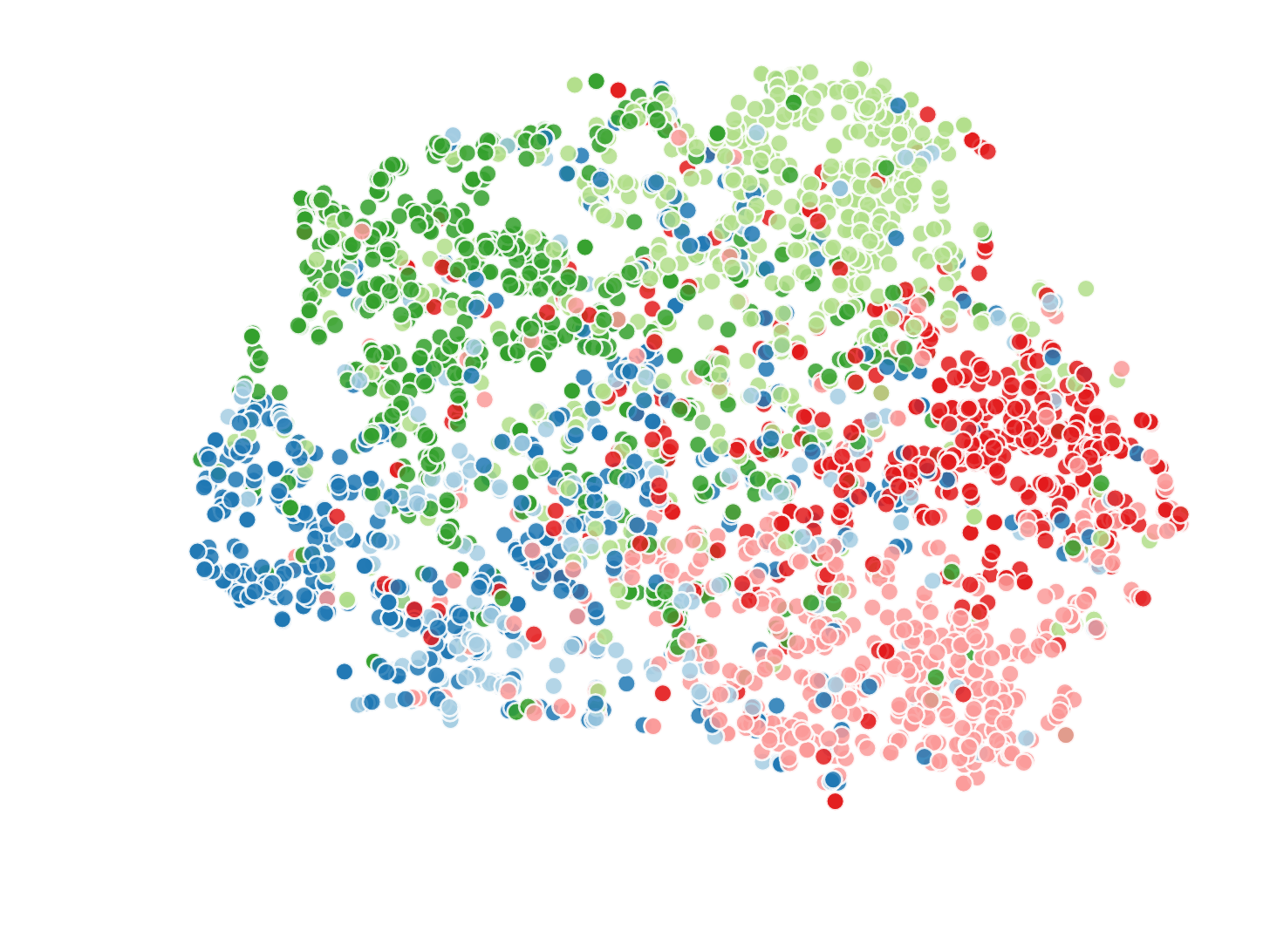}}
\subfigure[\GWNN on Citeseer] {\includegraphics[width=0.245\linewidth]{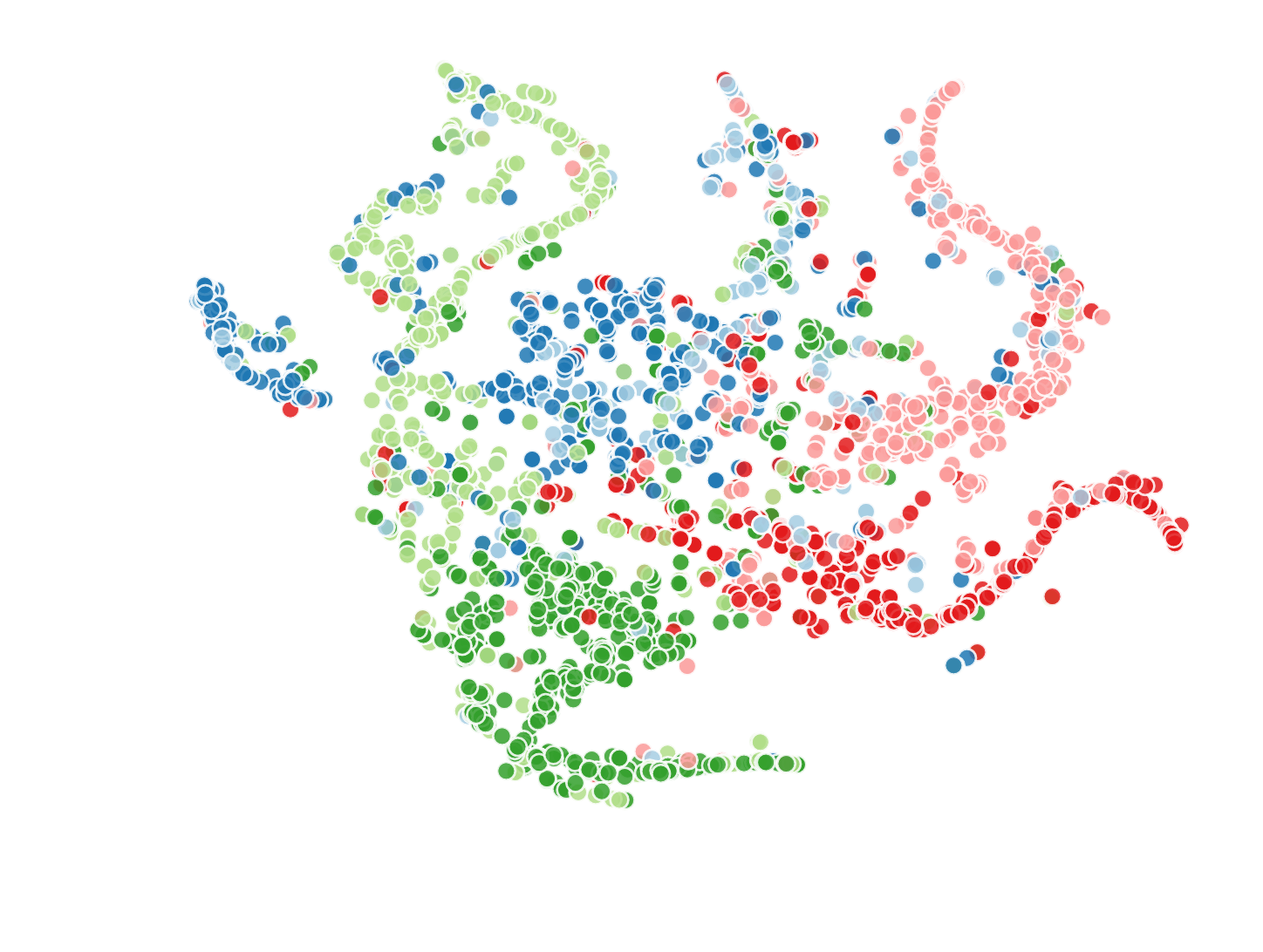}}
\subfigure[\GAT on Citeseer] {\includegraphics[width=0.245\linewidth]{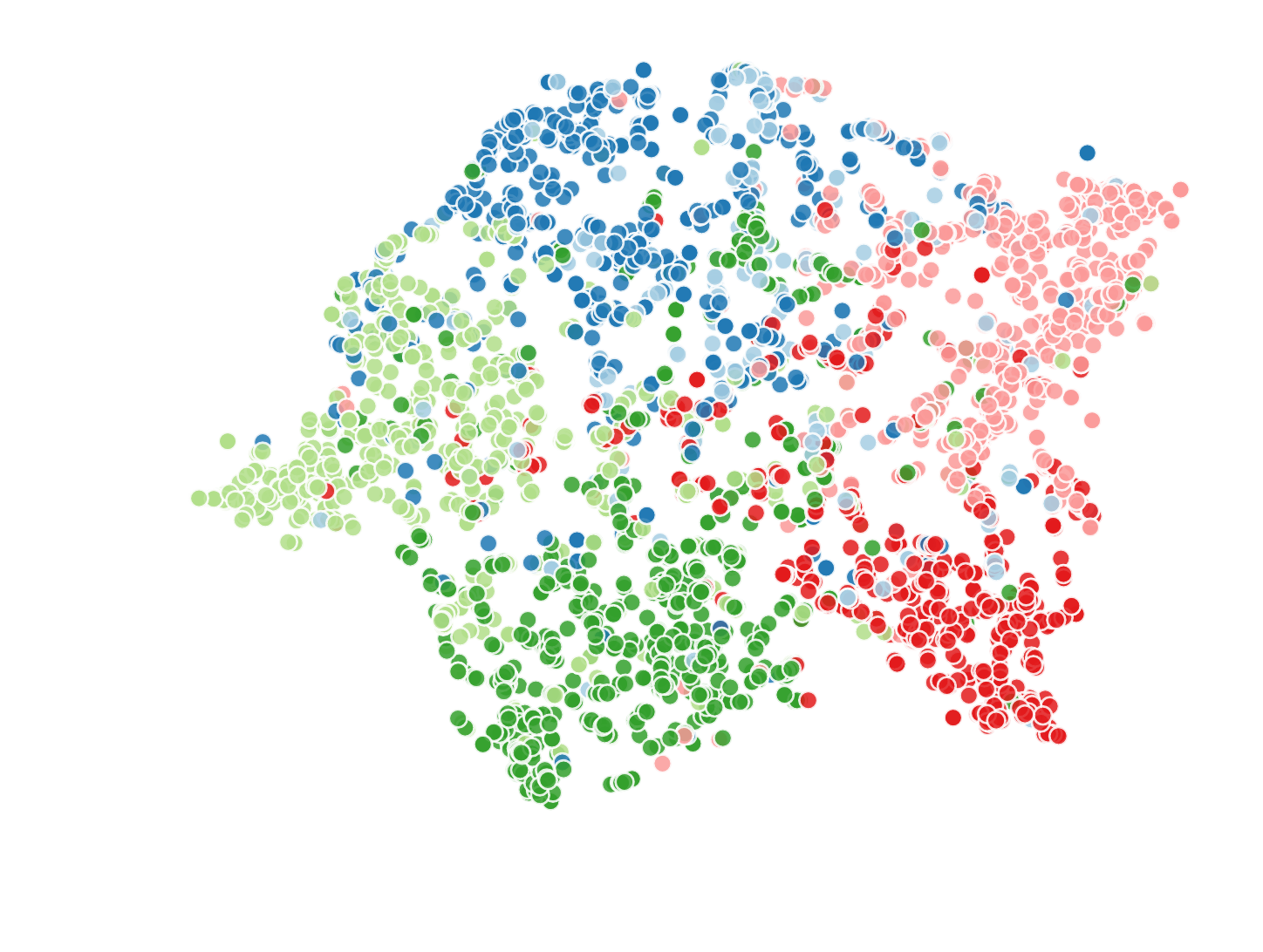}}
\subfigure [\SpGAT on Citeseer] {\includegraphics[width=0.245\linewidth]{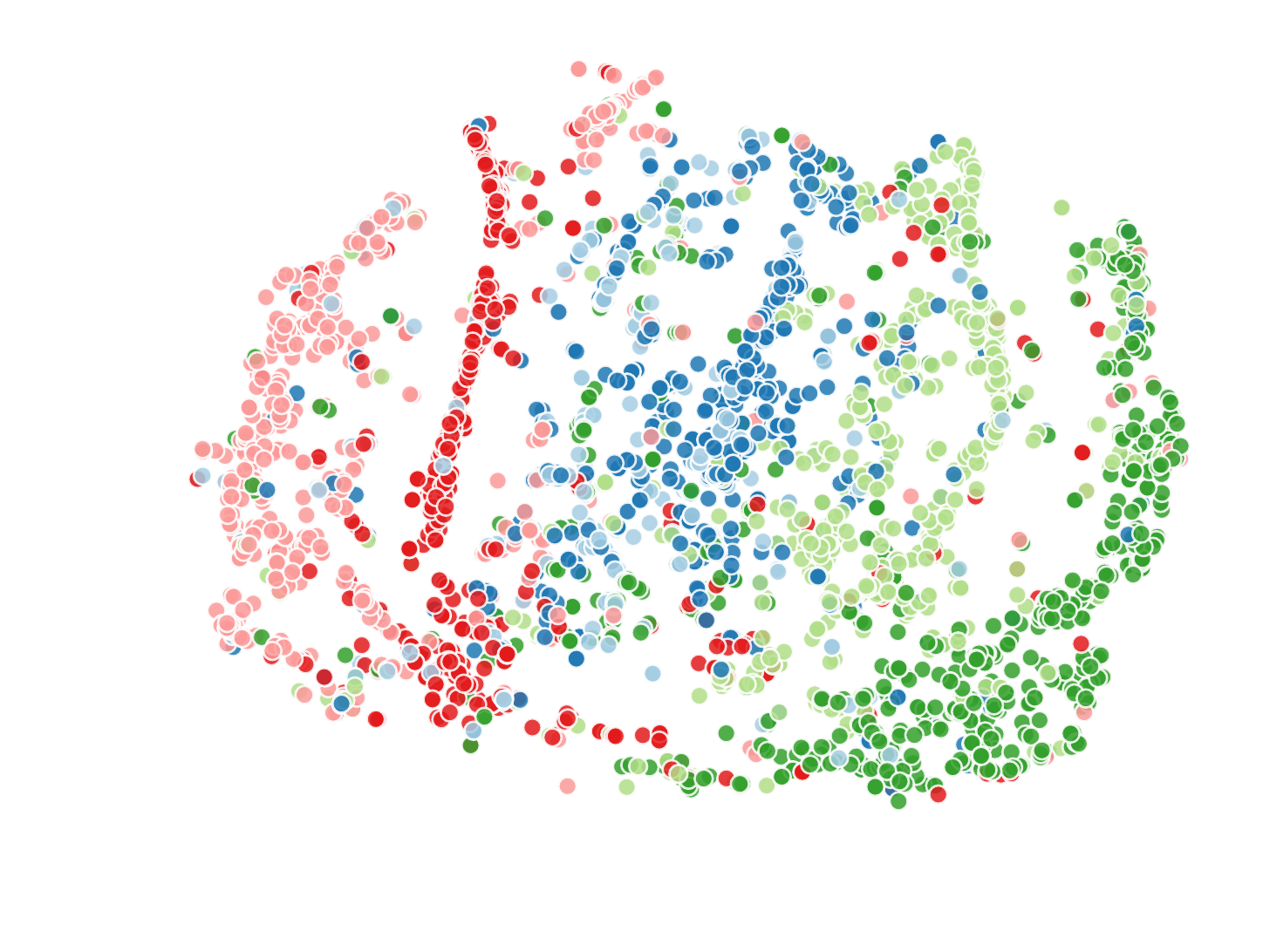}}
\vspace{7mm}
\subfigure [\VanillaGCN on Cora] {\includegraphics[width=0.245\linewidth]{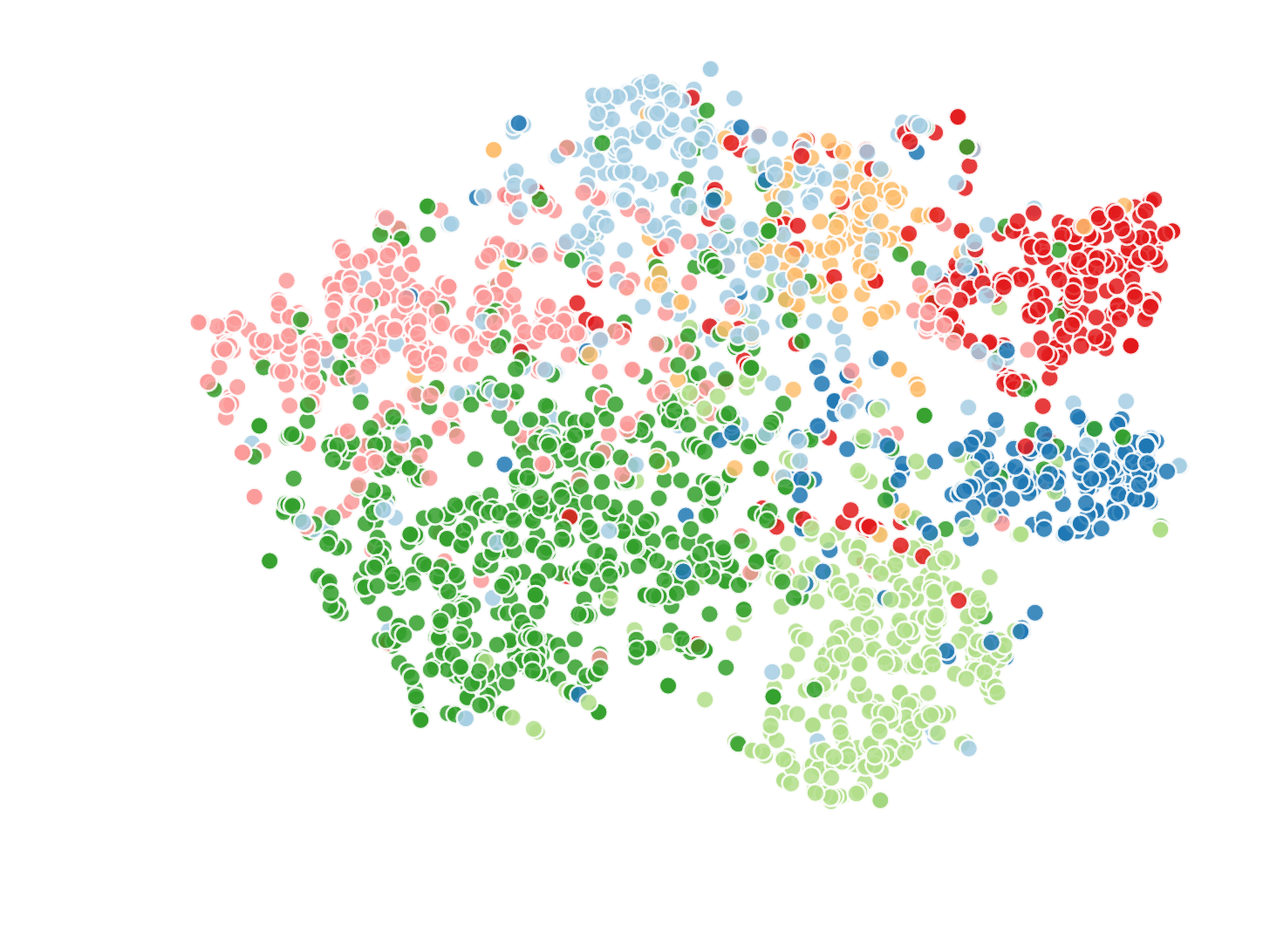}}
\subfigure[\GWNN on Cora] {\includegraphics[width=0.245\linewidth]{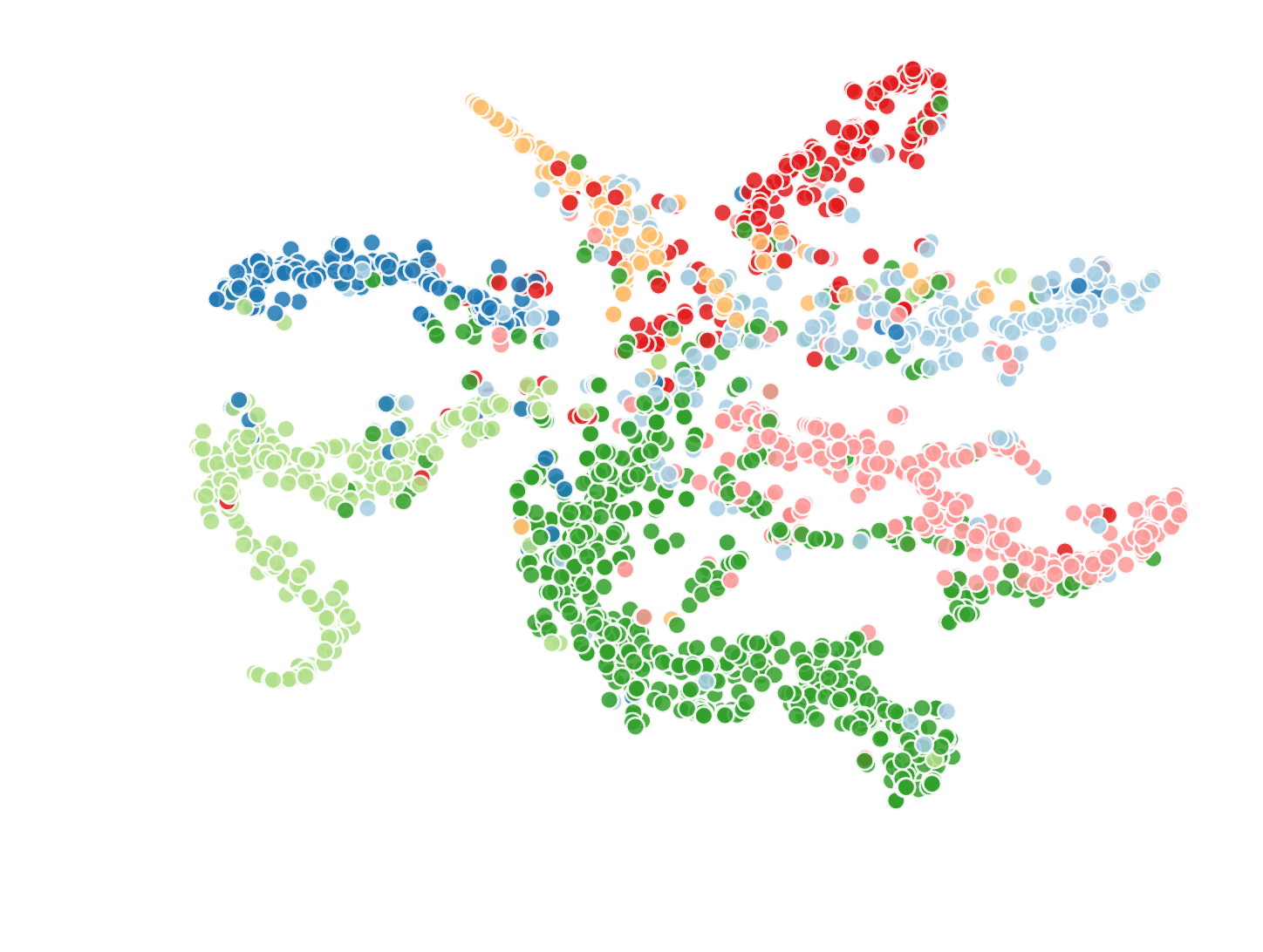}}
\subfigure[\GAT on Cora]
{\includegraphics[width=0.245\linewidth]{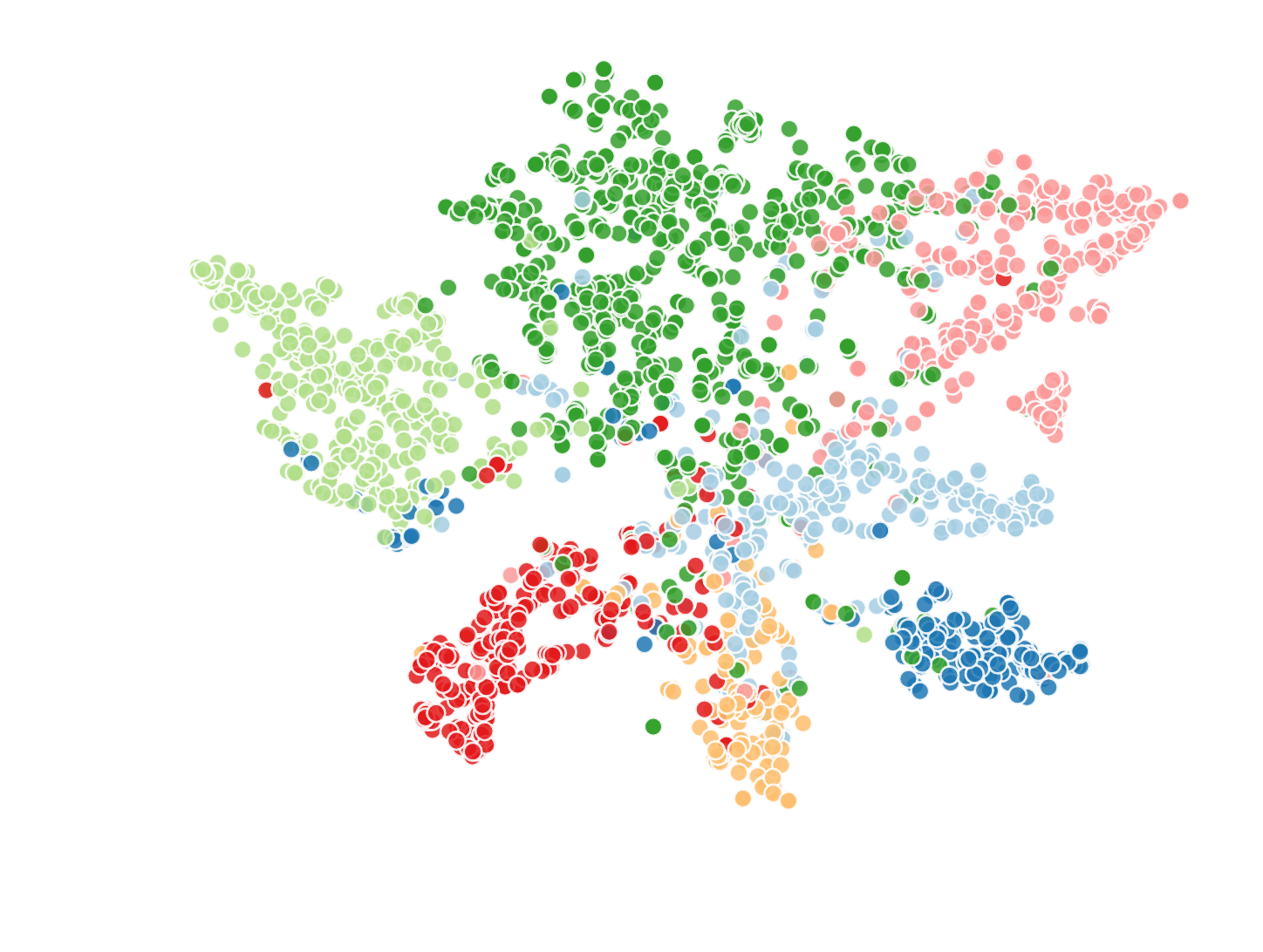}}
\subfigure [\SpGAT on Cora] {\includegraphics[width=0.245\linewidth]{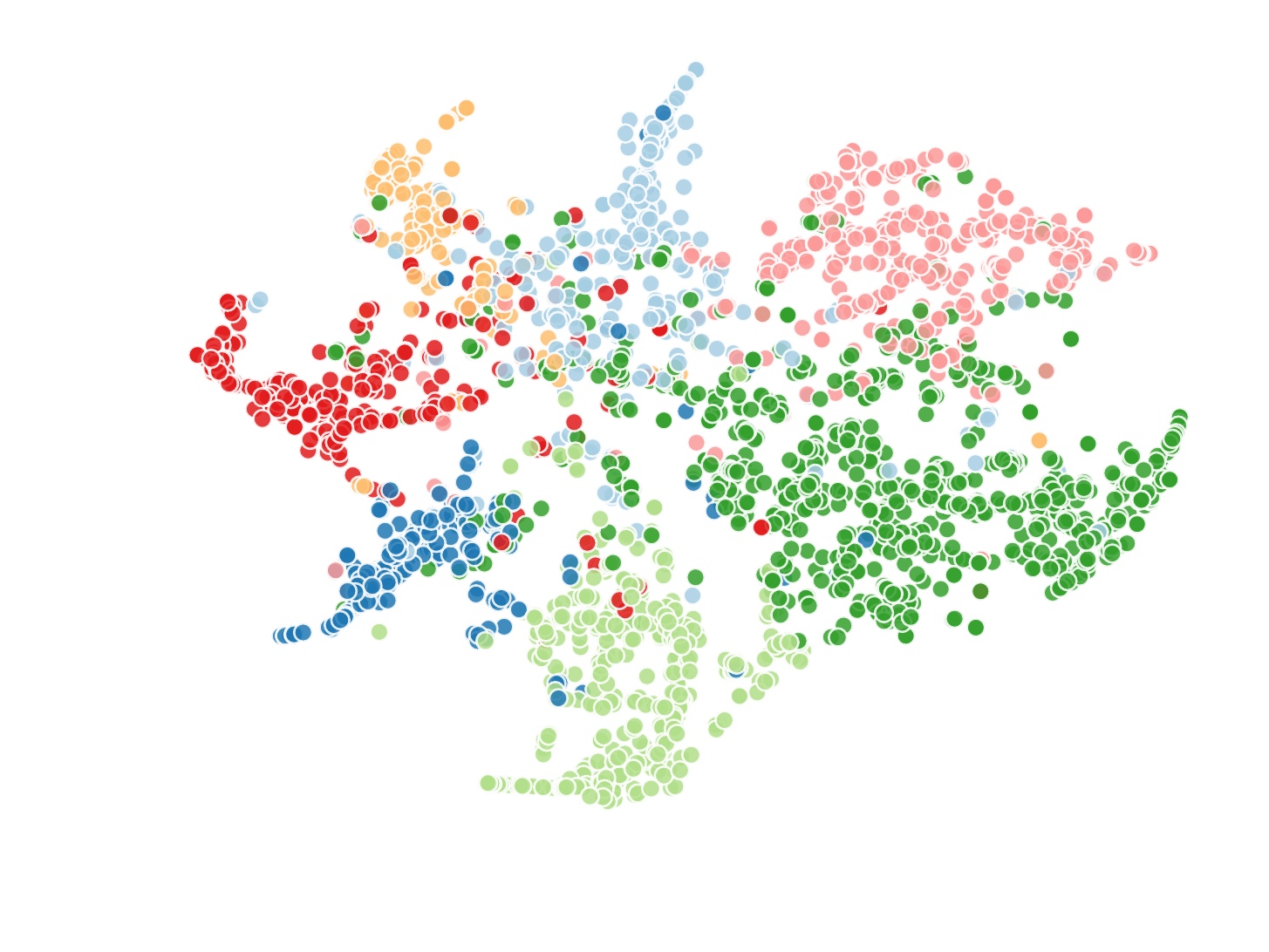}}
\subfigure [\VanillaGCN on Pubmed] {\includegraphics[width=0.245\linewidth]{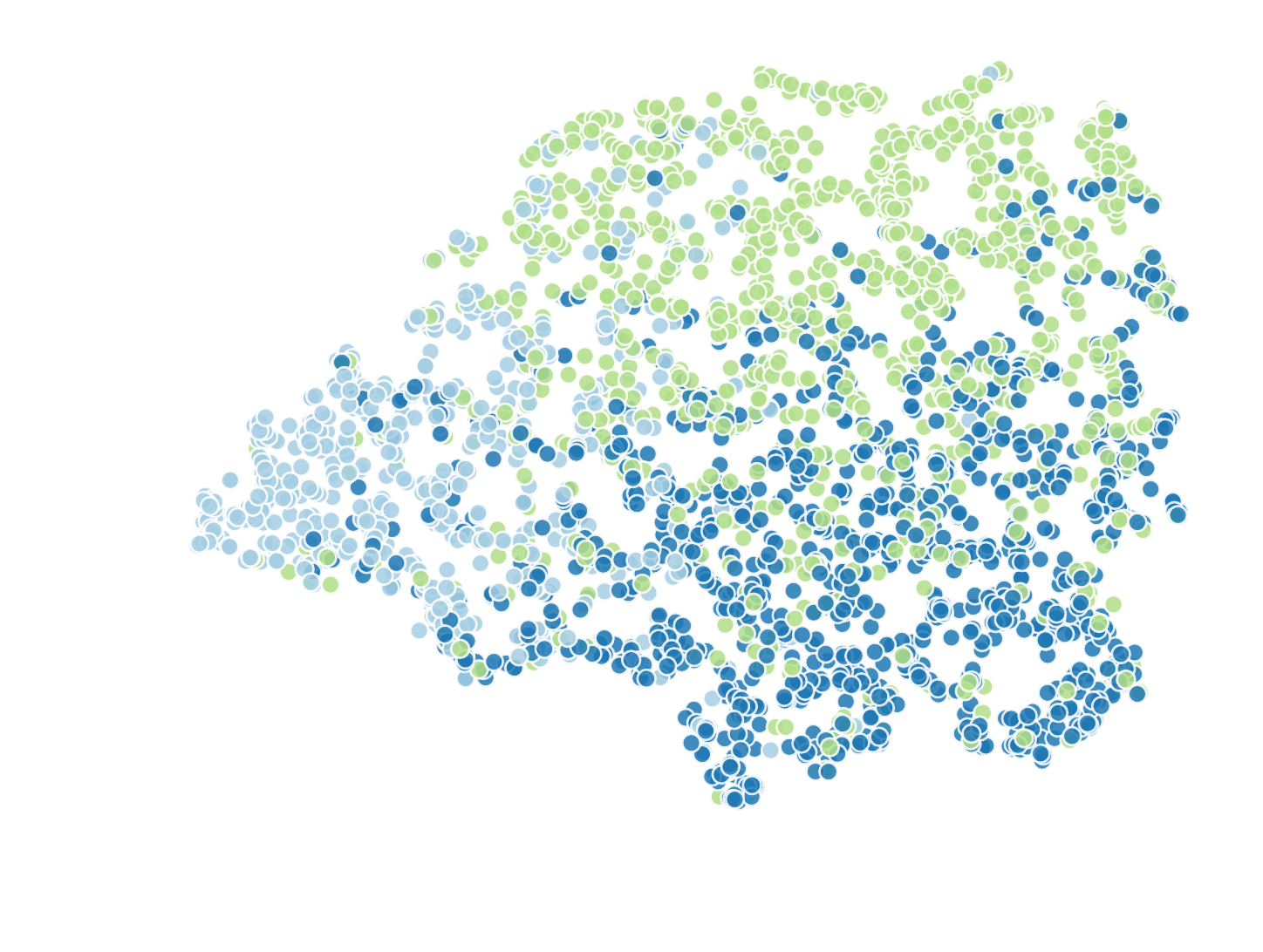}}
\subfigure[\GWNN on Pubmed] {\includegraphics[width=0.245\linewidth]{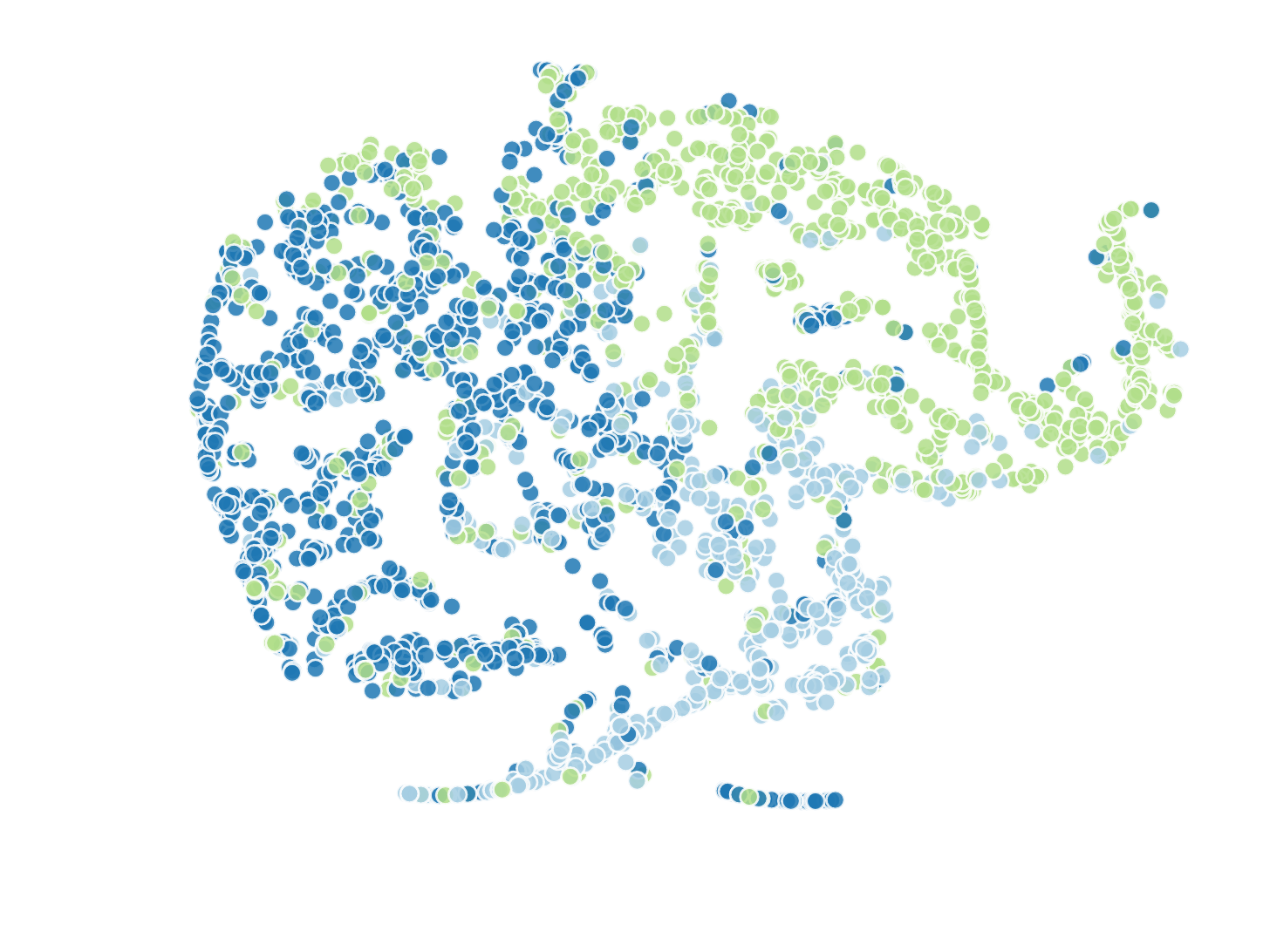}}
\subfigure[\GAT on Pubmed] {\includegraphics[width=0.245\linewidth]{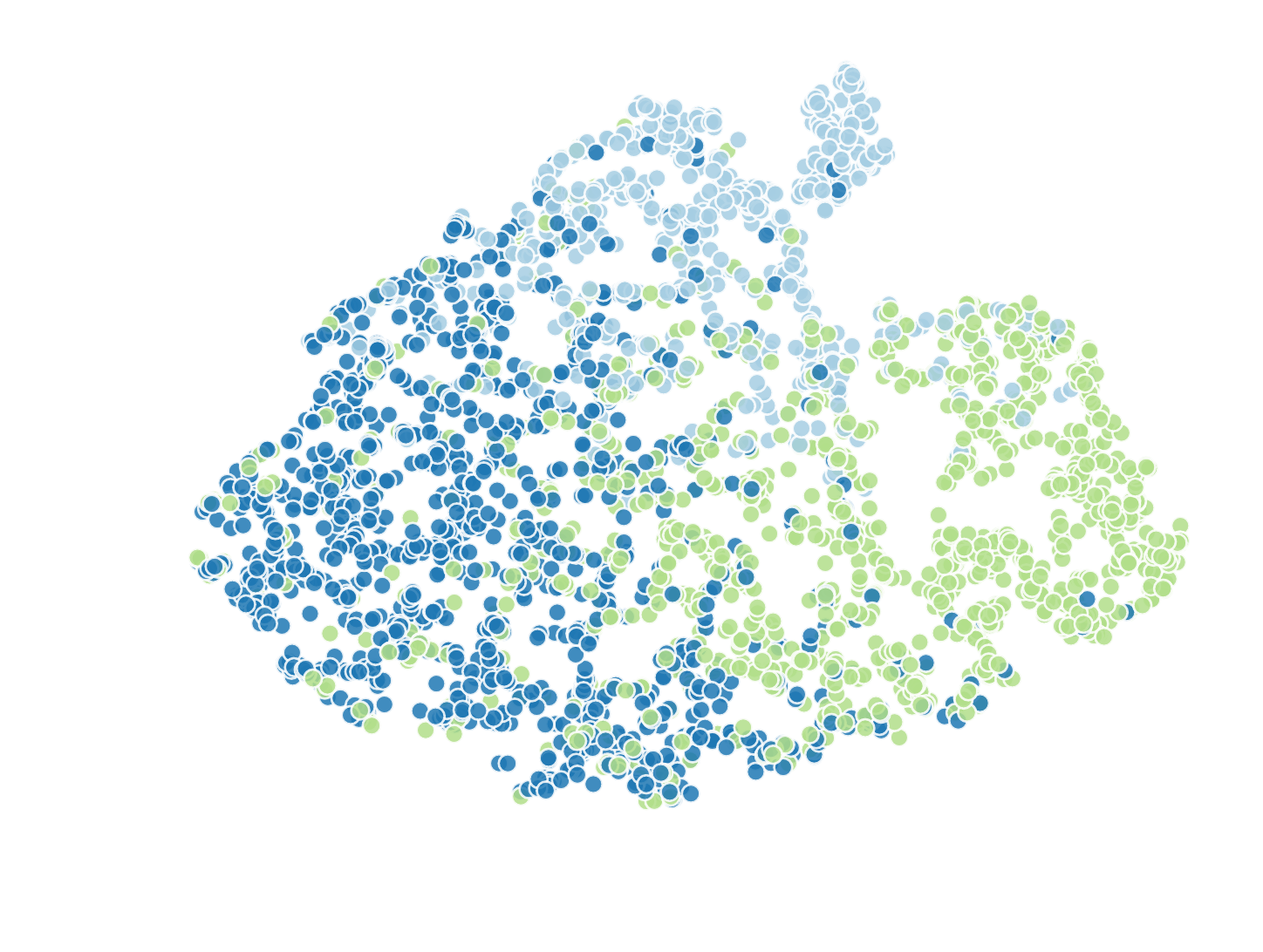}}
\subfigure [\SpGAT on Pubmed] {\includegraphics[width=0.245\linewidth]{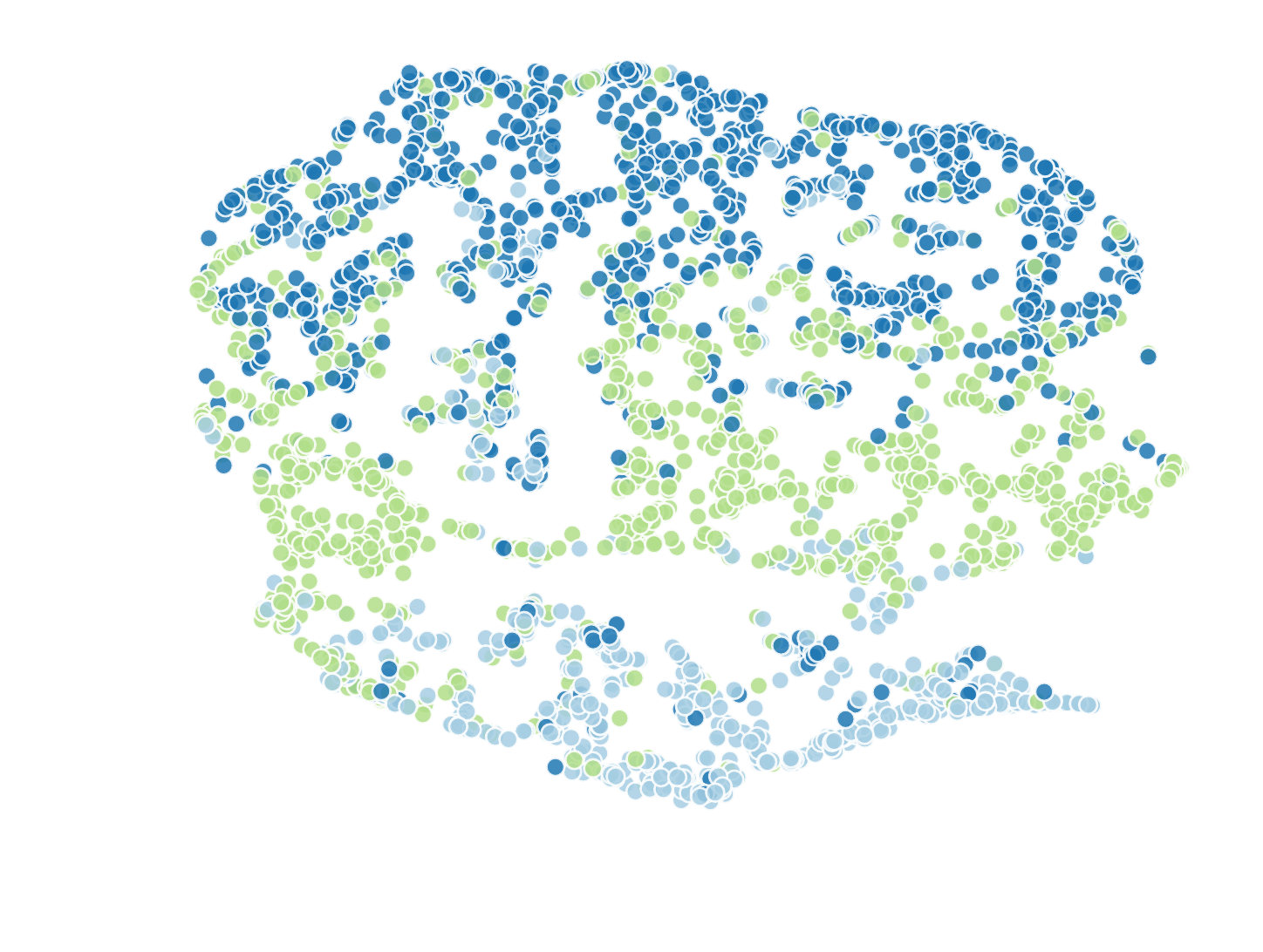}}
\vspace{-4mm}
\caption{The t-SNE visualization of \SpGAT comparing with other baselines on citation datasets. Each color corresponds to a different class that the embeddings belongs to.}
    \label{fig:tSNE}
\end{figure*}

\subsection{t-SNE Visualization of Learned Embeddings}
To evaluate the effectiveness of the learned features of \SpGAT qualitatively, we also depict the t-SNE visualization~\cite{maaten2008visualizing} of learned embeddings of \SpGAT on three citation datasets in Figure~\ref{fig:tSNE}.
The representation exhibits discernible clustering in the projected 2D space.
In Figure~\ref{fig:tSNE}, the color indicates the class label in each dataset. Compared with the other methods, the intersections of different classes in \SpGAT are more separated. It verifies the discriminative power of \SpGAT across the classes.

\section{Related Works}
\textbf{Spectral convolutional networks on graphs.} Existing methods of defining a convolutional operation on graphs can be broadly divided into two categories: spectral based and spatial based methods~\cite{zhang2020deep,guan2021autogl}. We focus on the spectral graph convolutions in this paper.
Spectral CNN~\cite{bruna2014spectral} first attempts to generalize CNNs to graphs based on the spectrum of the graph Laplacian and defines the convolutional kernel in the spectral domain.
ChebyNet~\cite{Defferrard2016ChebNet} introduces a fast localized convolutional filter on graphs via Chebyshev polynomial approximation.
Vanilla GCN~\cite{ICLR2017SemiGCN} further extends the spectral graph convolutions considering networks of significantly larger scale by several simplifications.
Lanczos algorithm is utilized in LanczosNet~\cite{liao2019lanczosnet} to construct low-rank approximations of the graph Laplacian for convolution.
\cite{ICLR2019GWNN} first attempts to construct graph neural networks with graph wavelets. 
SGC~\cite{wu2019simplifying} further reduces the complexity of Vanilla GCN by successively removing the non-linearities between consecutive layers. 
\cite{NIPS2019_Break} then generalizes the spectral graph convolution in block Krylov subspace forms to make use of multi-scale information.
Despite their effective performance, all these convolution theorem based methods lack the strategy to explicitly treat low- and high-frequency components with different importance.

\textbf{Space/spectrum-aware feature representation.}
In computer vision, \cite{chen2019drop} first defines space-aware feature representations based on scale-space theory and reduces spatial redundancy of vanilla CNN models by proposing the Octave Convolution (\OctConv) model.
To our knowledge, this is the first time that spectrum-aware feature representations are considered in irregular graph domain and established with graph convolutional neural networks.

\textbf{Spectral Graph Wavelets}
Theoretically, the lifting scheme is proposed for the construction of wavelets that can be adapted to irregular graphs in~\cite{sweldens1998lifting}.
\cite{Hammond2011Wavelets} defines wavelet transforms appropriate for graphs and describes a fast algorithm for computation via fast Chebyshev polynomial approximation.
For applications, \cite{tremblay2014graph} utilizes graph wavelets for multi-scale community mining and obtains a local view of the graph from each node. 
\cite{donnat2018learning} introduces the property of graph wavelets that describes information diffusion and learns structural node embeddings accordingly.
\cite{ICLR2019GWNN} first attempts to construct graph neural networks with graph wavelets. These works emphasize the local and sparse property of graph wavelets for Graph Signal Processing both theoretically and practically.

\end{document}


\title{Appendix: Spectral Graph Attention Network with Fast Eigen-approximation}

\author{Ben Trovato}
\authornote{Both authors contributed equally to this research.}
\email{trovato@corporation.com}
\orcid{1234-5678-9012}
\author{G.K.M. Tobin}
\authornotemark[1]
\email{webmaster@marysville-ohio.com}
\affiliation{%
  \institution{Institute for Clarity in Documentation}
  \streetaddress{P.O. Box 1212}
  \city{Dublin}
  \state{Ohio}
  \country{USA}
  \postcode{43017-6221}
}

\author{Charles Palmer}
\affiliation{%
  \institution{Palmer Research Laboratories}
  \streetaddress{8600 Datapoint Drive}
  \city{San Antonio}
  \state{Texas}
  \country{USA}
  \postcode{78229}}
\email{cpalmer@prl.com}

\renewcommand{\shortauthors}{Trovato and Tobin, et al.}




\maketitle

\section{Detailed Experimental Results}

\subsection{Datasets}
Joining the practice of previous works, we mainly focus on five node classification benchmark datasets under semi-supervised setting with different graph size and feature type.
(1) Three citation networks: Citeseer, Cora and Pubmed~\cite{Dataset2008Citeseer}, which aims to classify the research topics of papers. (2) A coauthor network: Coauthor CS which aims to predict the most active fields of study for each author from the KDD Cup 2016 challenge\footnote{https://kddcup2016.azurewebsites.net}. (3) A co-purchase network: Amazon Photo~\cite{mcauley2015image} which aims to predict the category of products from Amazon.
For the citation networks, we follow the public split setting provided by~\cite{yang2016revisiting}, that is, 20 labeled nodes per class in each dataset for training and 500 / 1000 labeled samples for validation / test respectively. For the other two datasets, we follow the splitting setting from~\cite{shchur2018pitfalls,chen2019measuring}.
Statistical overview of all datasets is given in Table~\ref{tab:datasets}. Label rate denotes the ratio of labeled nodes fetched in training process.

\begin{table}[htp]
\centering
\caption{\label{tab:datasets}The overview of dataset statistics.}
\resizebox{\columnwidth}{!}{%
\begin{tabular}{l r r r r r}
\toprule
\textbf{Dataset} & \textbf{Nodes} & \textbf{Edges} & \textbf{Classes} & \textbf{Features} & \textbf{Label rate}  \\[0.05em]\hline \\[-0.8em]
\textbf{Citeseer} & 3,327 & 4,732 & $6$ & 3,703 & $0.036$ \\
\textbf{Cora} & 2,708 & 5,429 & $7$ & 1,433 & $0.052$ \\
\textbf{Pubmed} & 19,717 & 44,338 & $3$ & 500 & $0.003$ \\
\textbf{Coauthor CS} & 18,333 & 81,894 & $15$ & 6,805 & $0.016$ \\
\textbf{Amazon Photo} & 7,487 & 11,9043 & $8$ & 745 & $0.021$ \\
\bottomrule
\end{tabular}
}
\end{table}

\section{Experimental Setup}
For all experiments, a 2-layer network of our model is constructed using TensorFlow~\cite{abadi2015tensorflow} with 64 hidden units. We train our model utilizing the Adam optimizer~\cite{kingma2014adam} with an initial learning rate $lr = 0.01$. We train the model using early stopping with a window size of 100. Most training process are stopped in less than 200 steps as expected. We initialize the weights matrix following~\cite{glorot2010understanding}, employ $5 \times 10^{-4}$ L2 regularization on weights and dropout input and hidden layers to prevent overfitting~\cite{srivastava2014dropout}.
For constructing wavelets $\psi_s(\lambda)$, we set $s=1$, $t=1\times 10^{-4}$ for \SpGAT and $M = 1$, $s = 2$ and $t=1\times 10^{-4}$ for \SpGATCheby on all datasets.
In addition, we employ the grid search to determine the best $d$ of low-frequency components and the impact of this parameter would be discussed in Section~\ref{sec.abstudy}. Two variants with \MEAN-pooling and \MAX-pooling are implemented to demonstrate the effectiveness of aggregation function in \SpGAT. Without other specification in our experimental setup, we use the \MAX-pooling function in \SpGAT.

\begin{figure}[htb]
\centering
\includegraphics [width=0.7\columnwidth]{figs/F1_comparison.pdf}
\vspace{-2ex}
\caption{Inductive node classification task on PPI.}
\vspace{-2mm}
\label{fig:F1_comparison}
\end{figure}

\subsection{Performance under Inductive Node Classification}
Aside from the transductive tasks we analyzed above, we further empirically conduct the inductive node classification task on Protein-Protein Interaction (PPI) dataset. 
PPI models the interactions between proteins using a graph, with nodes being proteins and edges being interactions.
Each protein has at most 121 labels and be associated with additional 50-dimensional features. The train/valid/test split is consistent with \GraphSAGE~\cite{Hamilton2017Inductive}. We compare the performance with three baselines: \VanillaGCN~\cite{ICLR2017SemiGCN},
\GraphSAGE~\cite{Hamilton2017Inductive} and \GAT~\cite{velickovic2018gat}.
As shown in Figure~\ref{fig:F1_comparison}, \SpGATCheby is capable of handling with inductive task with comparable performance against \GAT, which implies the transferability of spectral attention weights among graphs. The in-depth analysis of this transferability would be an interesting future work. Please note that \SpGAT is motivated in the transductive setting and needs a full eigen-decomposition, which makes it not suitable for the inductive setting and is not shown in Figure~\ref{fig:F1_comparison}.

\section{Full Ablation Studies}\label{sec.abstudy}

\subsection{The Learned Attention on Low- and High-frequency Components}
In this Section, we also show how the learned attentions of \SpGAT w.r.t the best proportion for Citeseer, Cora and Pubmed which are demonstrated in Table~\ref{tab:learned alpha weights}. Interestingly, despite the small proportion, the attention weight of low-frequency components learned by \SpGAT is much larger than that of high-frequency components in each layer consistently. Hence, \SpGAT is successfully to capture the importance of low- and high-frequency components of graphs in the spectral domain. Moreover, as pointed out by~\cite{donnat2018learning,maehara2019revisiting}, the low-frequency components in graphs usually indicate smooth varying signals which can reflect the locality property in graphs. It implies that the local structural information is important for these datasets. This may explain why \GAT also gains good performance on these datasets.

\begin{table}
\caption{Learned attention weights $\alpha_{L}$ and $\alpha_{H}$ of  \SpGAT for low- and high-frequency w.r.t the best proportion of low frequency components $d$ (number followed after the name of datasets). \label{tab:learned alpha weights}}
\vspace{-2mm}
\centering
\resizebox{0.49\textwidth}{!}{%
\begin{tabular}{@{} lcc|cc|cc@{}}
\toprule
    Dataset & \multicolumn{2}{c|}{Citeseer ($15\%$)} & \multicolumn{2}{c|}{Cora ($5\%$)} & \multicolumn{2}{c}{Pubmed ($10\%$)}  \\
\midrule
Attention filter weights & $\alpha_{L}$  & $\alpha_{H}$   & $\alpha_{L}$  & $\alpha_{H}$  & $\alpha_{L}$  & $\alpha_{H}$   \\
\midrule
Learned value (first layer) & $\textbf{0.84}$ &  $0.16$ &   $\textbf{0.72}$ & $0.23$ & $\textbf{0.86}$ & $0.14$ \\
\midrule
Learned value (second layer) & $\textbf{0.94}$ &  $0.06$ &  $\textbf{0.93}$ & $0.07$ & $$\textbf{0.93}$$ & $0.07$ \\
\bottomrule
\end{tabular}
}
\end{table}

\begin{table}[htb]
\caption{The results of ablation study on low- and high-frequency components. \label{tab:test only with L/H}}
\vspace{-2mm}
\centering
\resizebox{\columnwidth}{!}{
\begin{tabular}{@{} lccc@{}}
\toprule
    Methods & {Citeseer ($15\%$)} & {Cora ($5\%$)} & {Pubmed ($10\%$)}  \\
\midrule
with low-frequency & $57.7$ & $66.8$ & $76.7$ \\
\midrule
with high-frequency & $70.9$ & $82.4$ & $80.4$ \\
\midrule
\SpGAT & $\textbf{73.9}$ & $\textbf{84.2}$ & $\textbf{80.8}$ \\
\bottomrule
\end{tabular}
}
\vspace{-1mm}
\end{table}

\subsection{Only Low- and High-frequency Components}
To further elaborate the importance of low- and high-frequency components in \SpGAT, we conduct the ablation study on the classification results by testing only with low- or high-frequency components w.r.t the best proportion. Specially, we manually set $\alpha_{L}$ or $\alpha_{H}$ to 0 during testing stage to observe how the learned low- and high-frequency components in graphs affect the classification accuracy. From Table~\ref{tab:test only with L/H}, we can observe that: 
\begin{itemize}
    \item  Both low- and high-frequency components are essential for the model. Since removing any components downgrade the over performance.
    \item \SpGAT with very small proportion (5\% - 15\%) of low-frequency components can achieve the comparable results to those obtained by full \SpGAT. It reads that the low-frequency components contain more information that can contribute to the feature representation learned from the model.
\end{itemize}

\begin{table}[htb]
\caption{Demonstration of the effectiveness the simplicity of the designed mechanism. \label{tab:simplicity}}
\vspace{-2mm}
\centering
\resizebox{0.85\columnwidth}{!}{
\begin{tabular}{@{} lccccc@{}}
\toprule
Methods & \GCNII & \SpectralCNN & \SpGATNA & \SpGATF-\MAX & \SpGAT-\MAX  \\
\midrule
Cora & $\textbf{84.5}$ & $73.3$ & $74.9$ & $80.9$ & $\textbf{84.4}$ \\
\bottomrule
\end{tabular}
}
\vspace{-1mm}
\end{table}

\begin{table}[htb]
\caption{Parameters reduction comparison with baselines. (The best results are in boldface). \label{tab:parameters}}
\centering
\begin{tabular}{@{} lcccc@{}}
\toprule
Methods & \ChebyNet & \GWNN & \GCNII & \SpGAT  \\
\midrule
Parameters & $46,080$ & $23,048$ & $34,719$ & $\textbf{12,632}$ \\
\bottomrule
\end{tabular}
\end{table}

\begin{table}[htb]
\caption{Additional accuracy results on the comparison with \GCNII. (The best results are in boldface). \label{tab:comparable GCNII}}
\centering
\resizebox{0.33\textwidth}{!}{%
\begin{tabular}{@{} lccc@{}}
\toprule
Dataset & Citeseer & Cora & Pubmed \\
\midrule
\VanillaGCN & $70.3$   & $81.5$  & $79.0$ \\
\midrule
\GCNII & $73.4$ &  $\textbf{85.5}$ & $80.2$ \\
\midrule
\SpGAT & $\textbf{73.9}$ &  $84.2$ & $\textbf{80.8}$ \\
\bottomrule
\end{tabular}
}
\end{table}

\begin{table*}[!t]
\caption{The mean Silhouette Coefficient of learned samples. Larger is better. \label{tab:Silhouette Coefficient}}
\centering
\resizebox{\textwidth}{!}{%
\begin{tabular}{@{} lcccc|cccc|cccc@{}}
\toprule
    Dataset & \multicolumn{4}{c|}{Citeseer} & \multicolumn{4}{c|}{Cora} & \multicolumn{4}{c}{Pubmed}  \\
\midrule
Model & \VanillaGCN & \GWNN  & \GAT & \SpGAT  & \VanillaGCN & \GWNN  & \GAT & \SpGAT   & \VanillaGCN & \GWNN  & \GAT & \SpGAT   \\
\midrule
Silhouette Coefficient & $0.038$  & $0.050$ & $0.102$ & \textbf{0.130} & $0.119$ & $0.153$ & $0.230$ & \textbf{0.243} &  $0.110$ & $0.130$ & $0.168$ & \textbf{0.195} \\
\bottomrule
\end{tabular}
}
\end{table*}

\begin{figure*}[!t]
\centering
\subfigure [\VanillaGCN on Citeseer] {\includegraphics[width=0.245\linewidth]{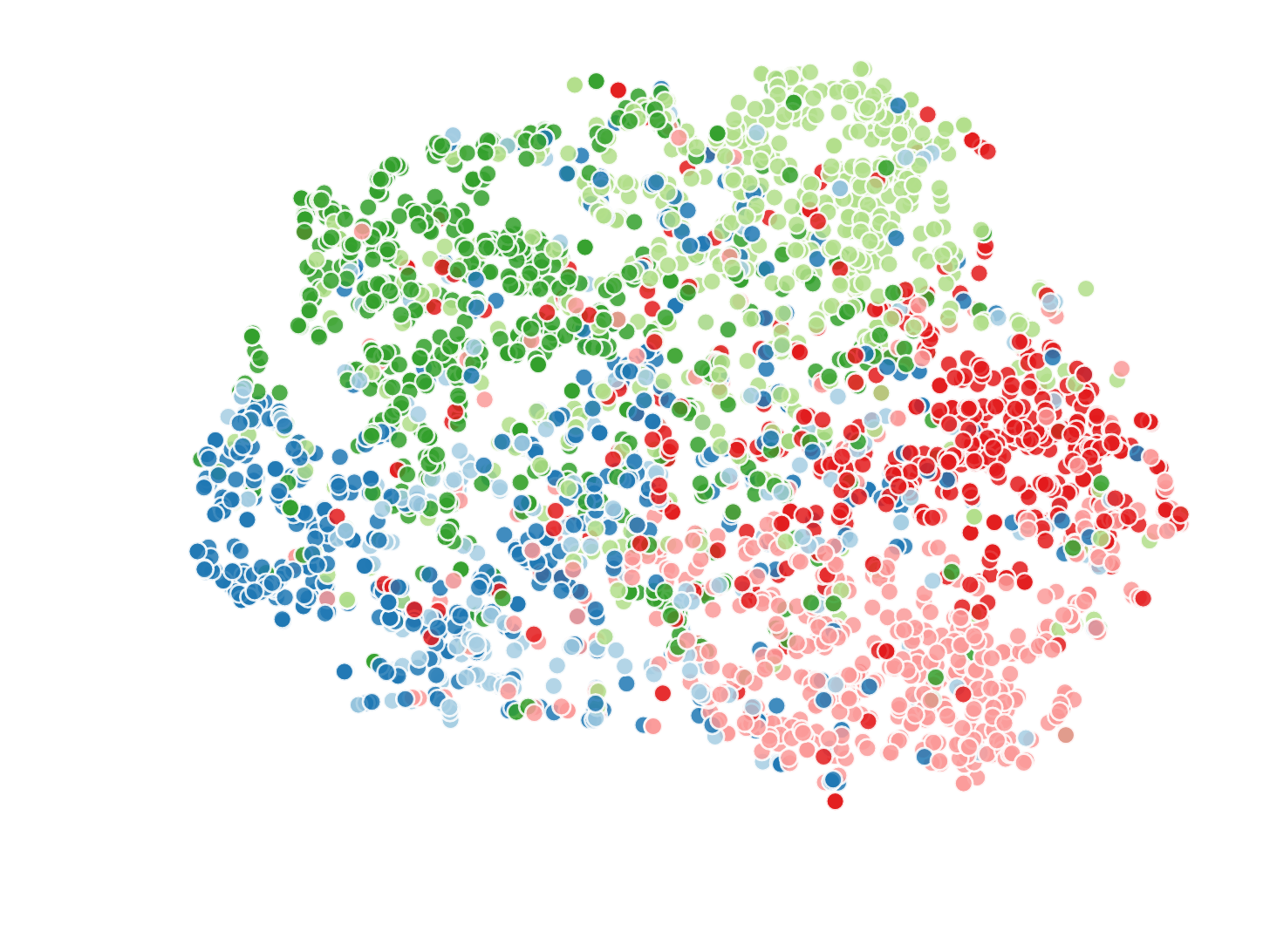}}
\subfigure[\GWNN on Citeseer] {\includegraphics[width=0.245\linewidth]{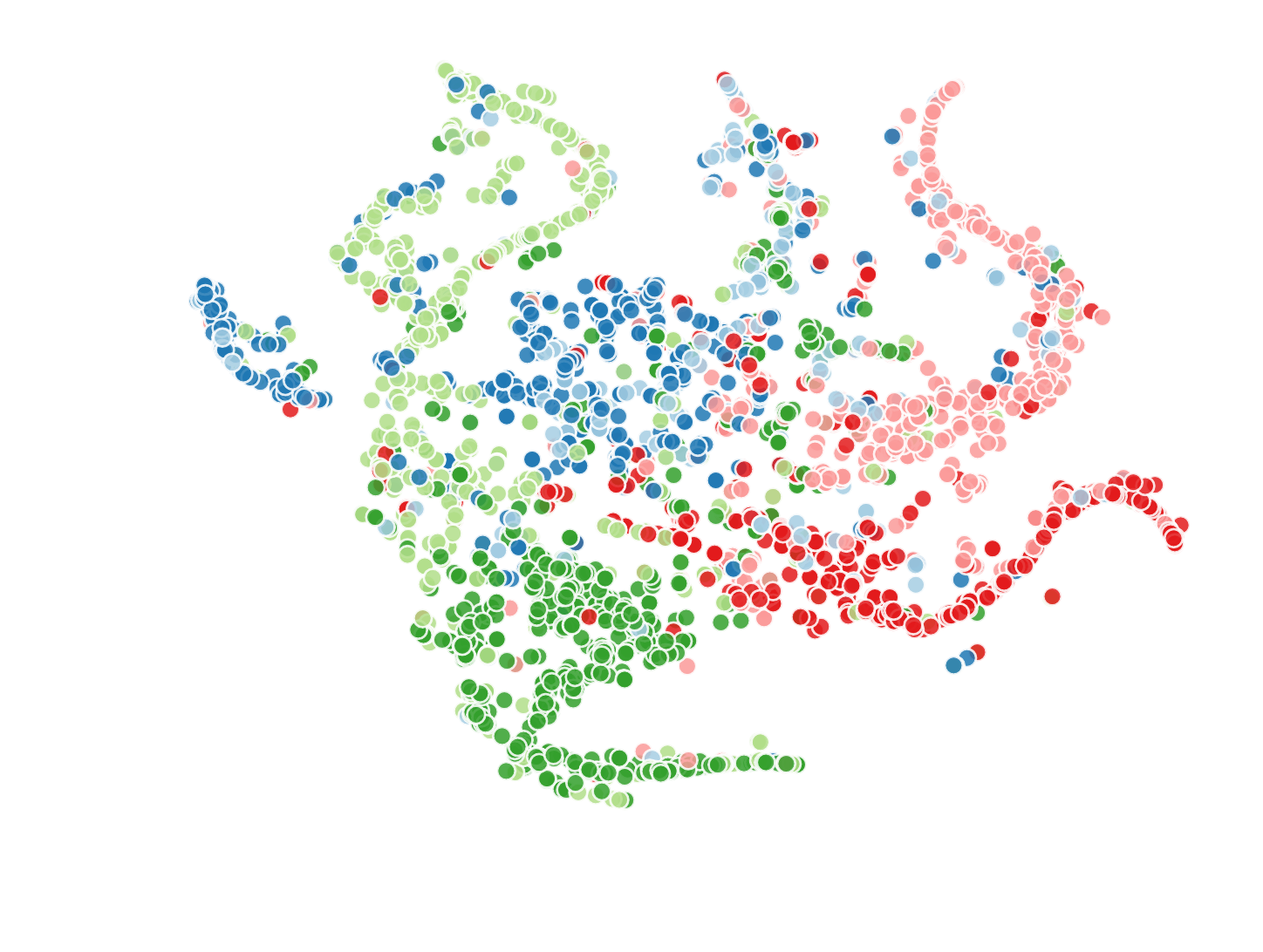}}
\subfigure[\GAT on Citeseer] {\includegraphics[width=0.245\linewidth]{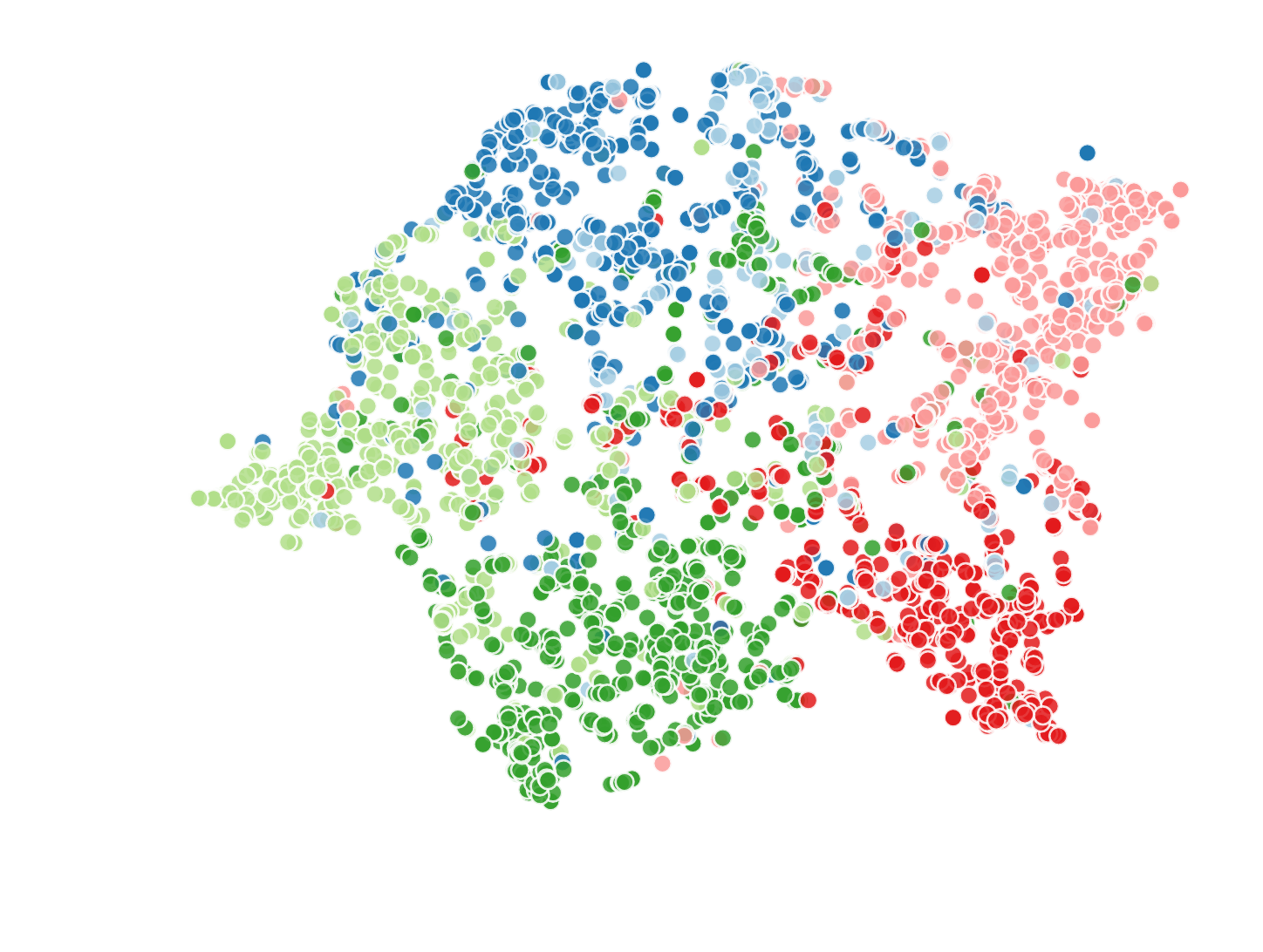}}
\subfigure [\SpGAT on Citeseer] {\includegraphics[width=0.245\linewidth]{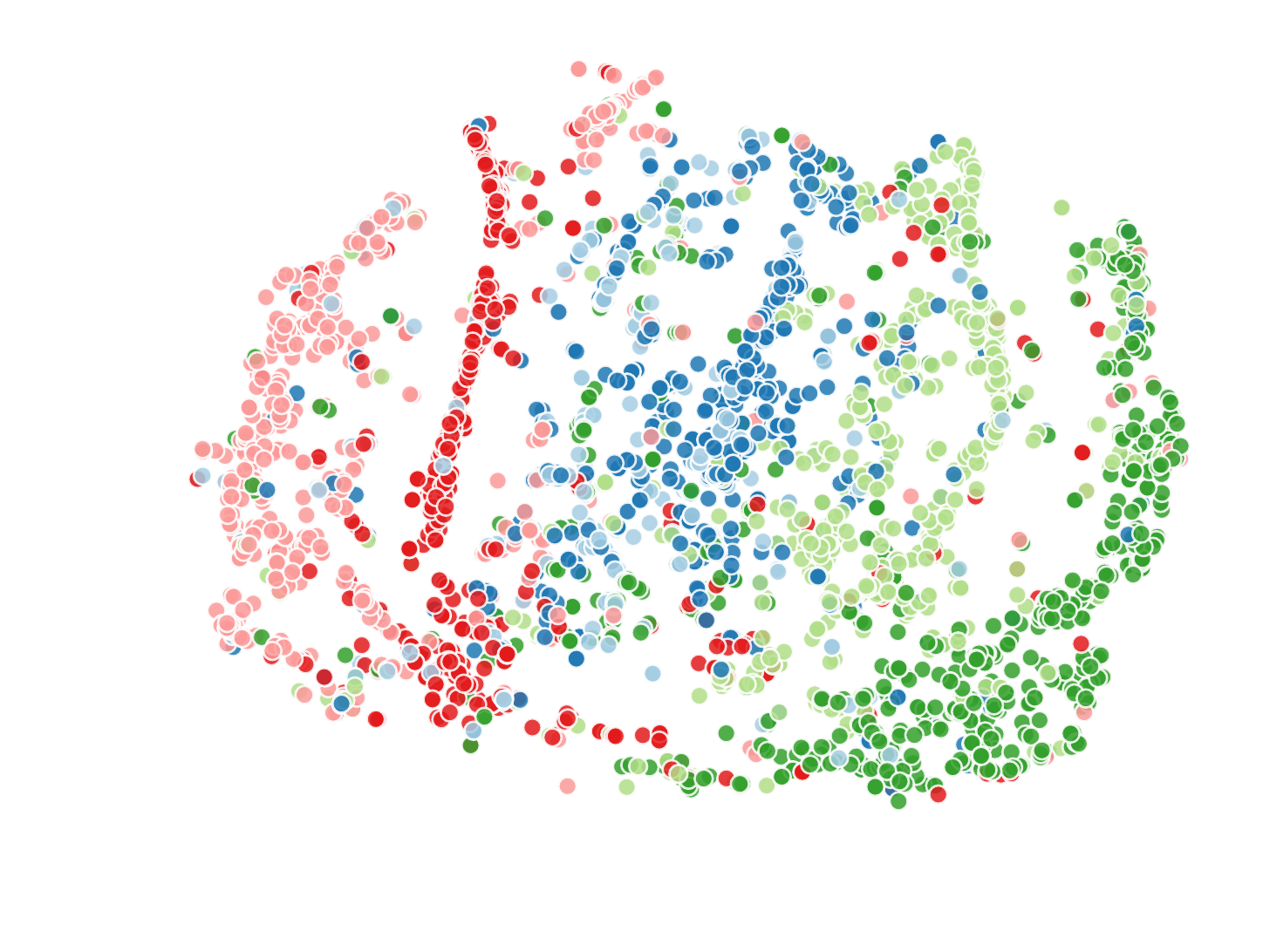}}
\vspace{7mm}
\subfigure [\VanillaGCN on Cora] {\includegraphics[width=0.245\linewidth]{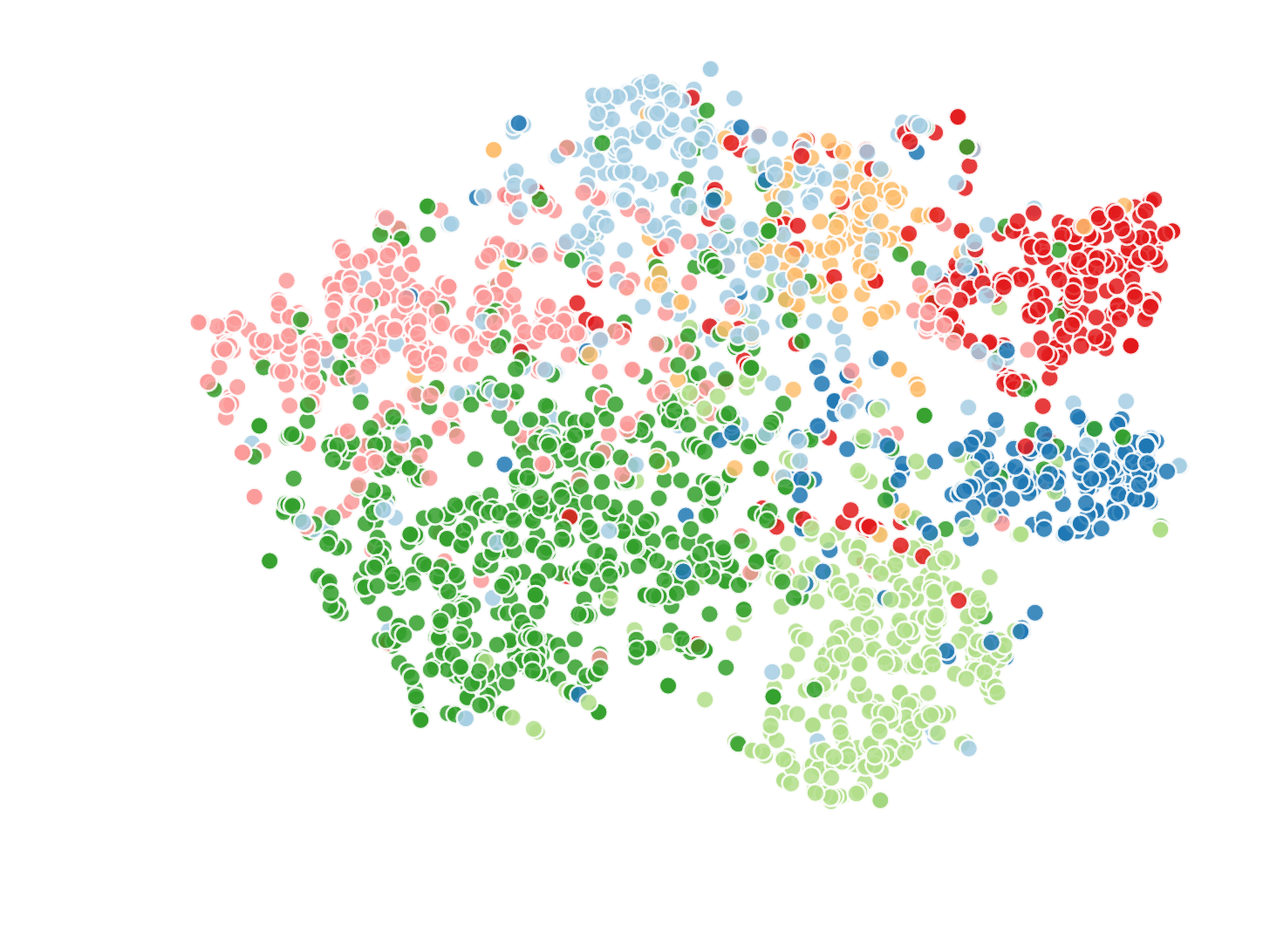}}
\subfigure[\GWNN on Cora] {\includegraphics[width=0.245\linewidth]{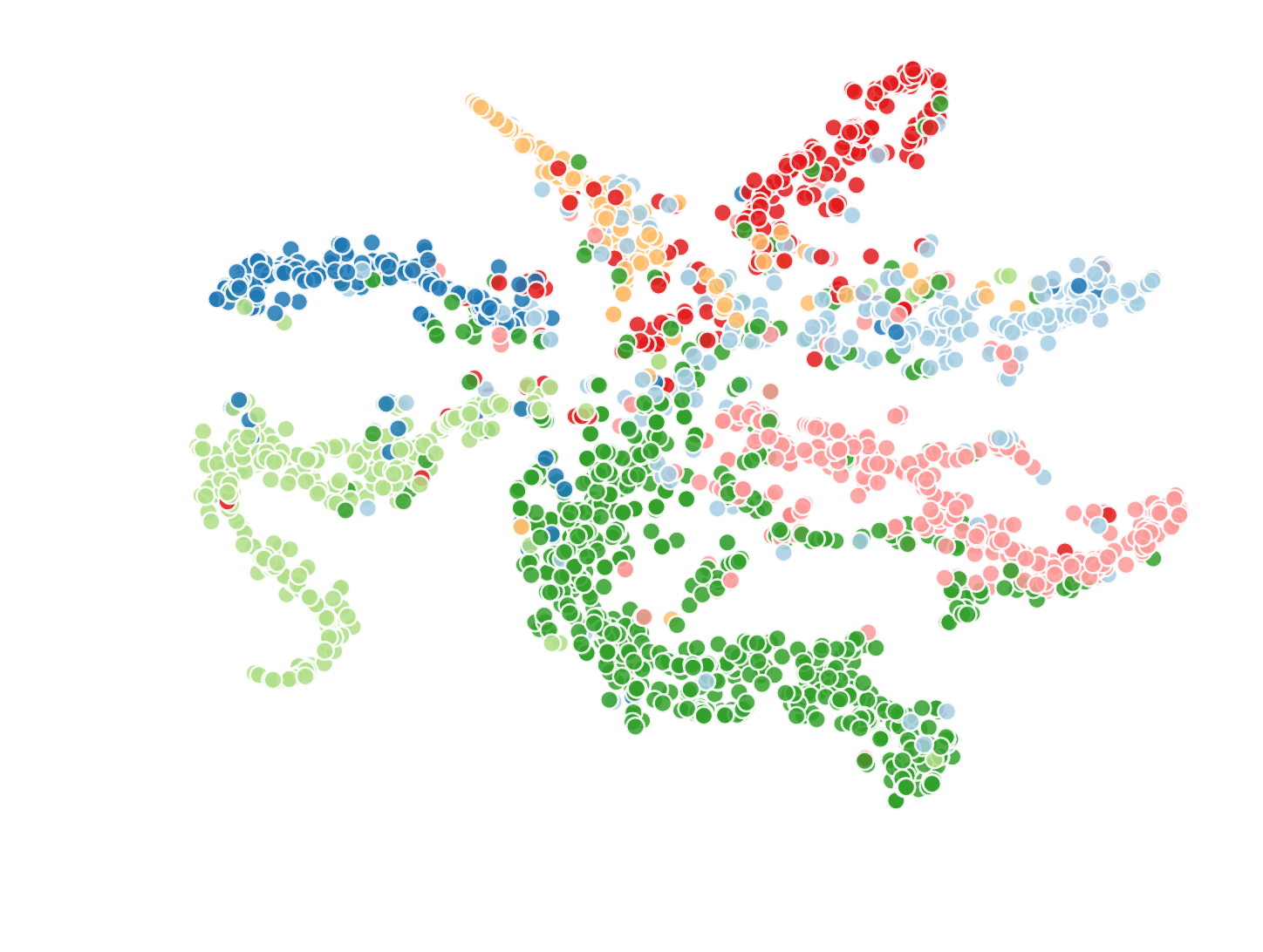}}
\subfigure[\GAT on Cora]
{\includegraphics[width=0.245\linewidth]{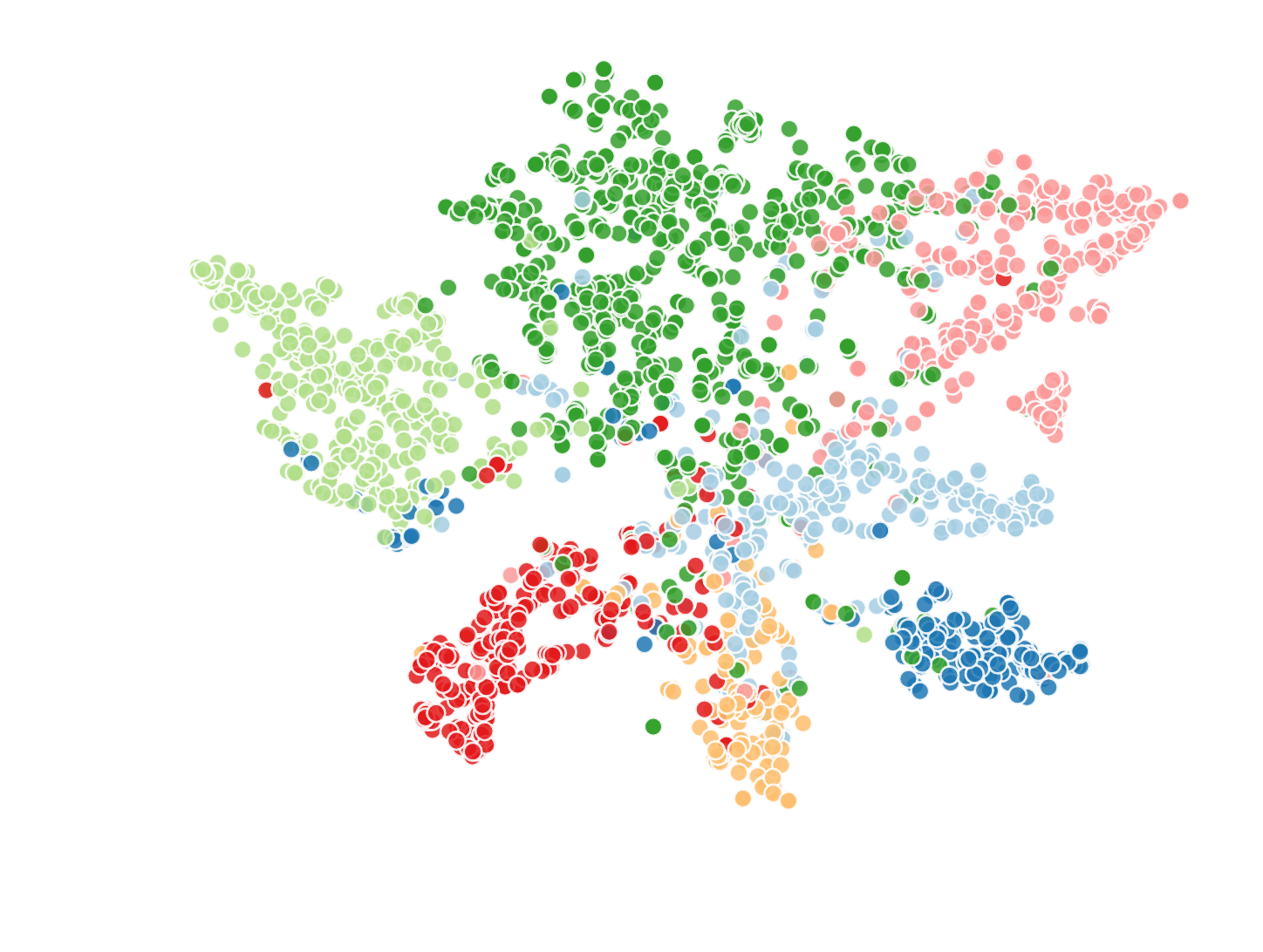}}
\subfigure [\SpGAT on Cora] {\includegraphics[width=0.245\linewidth]{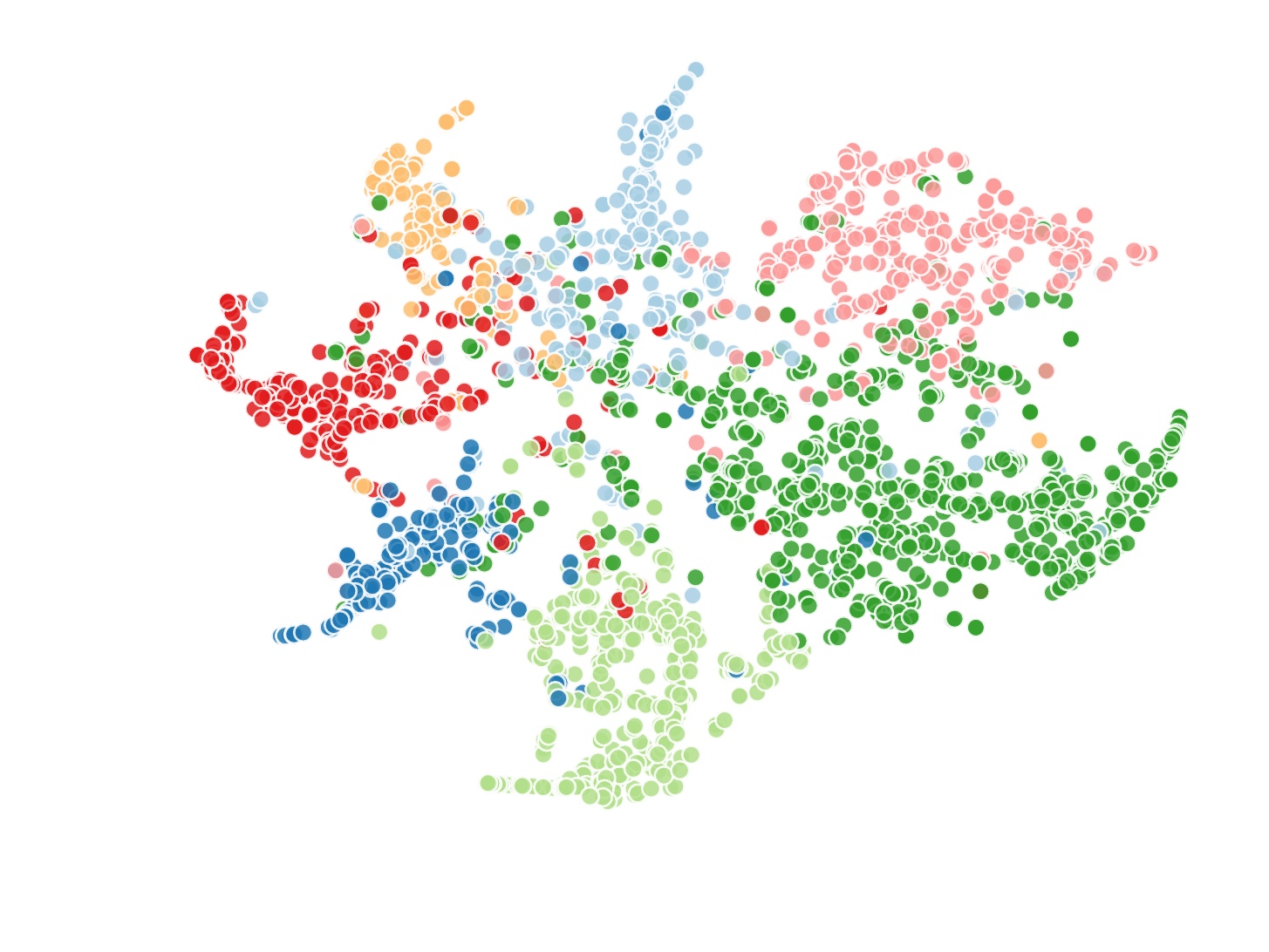}}
\subfigure [\VanillaGCN on Pubmed] {\includegraphics[width=0.245\linewidth]{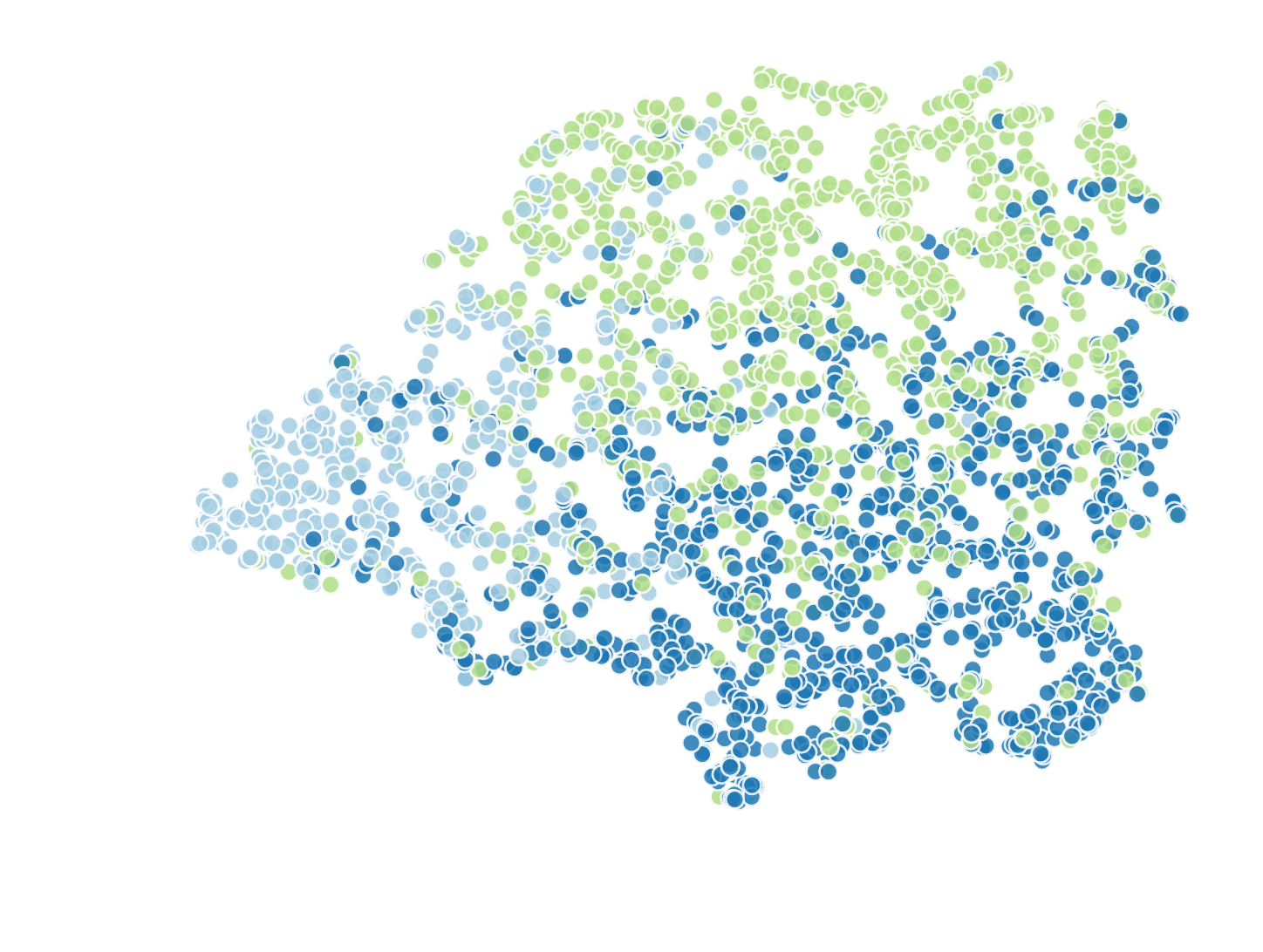}}
\subfigure[\GWNN on Pubmed] {\includegraphics[width=0.245\linewidth]{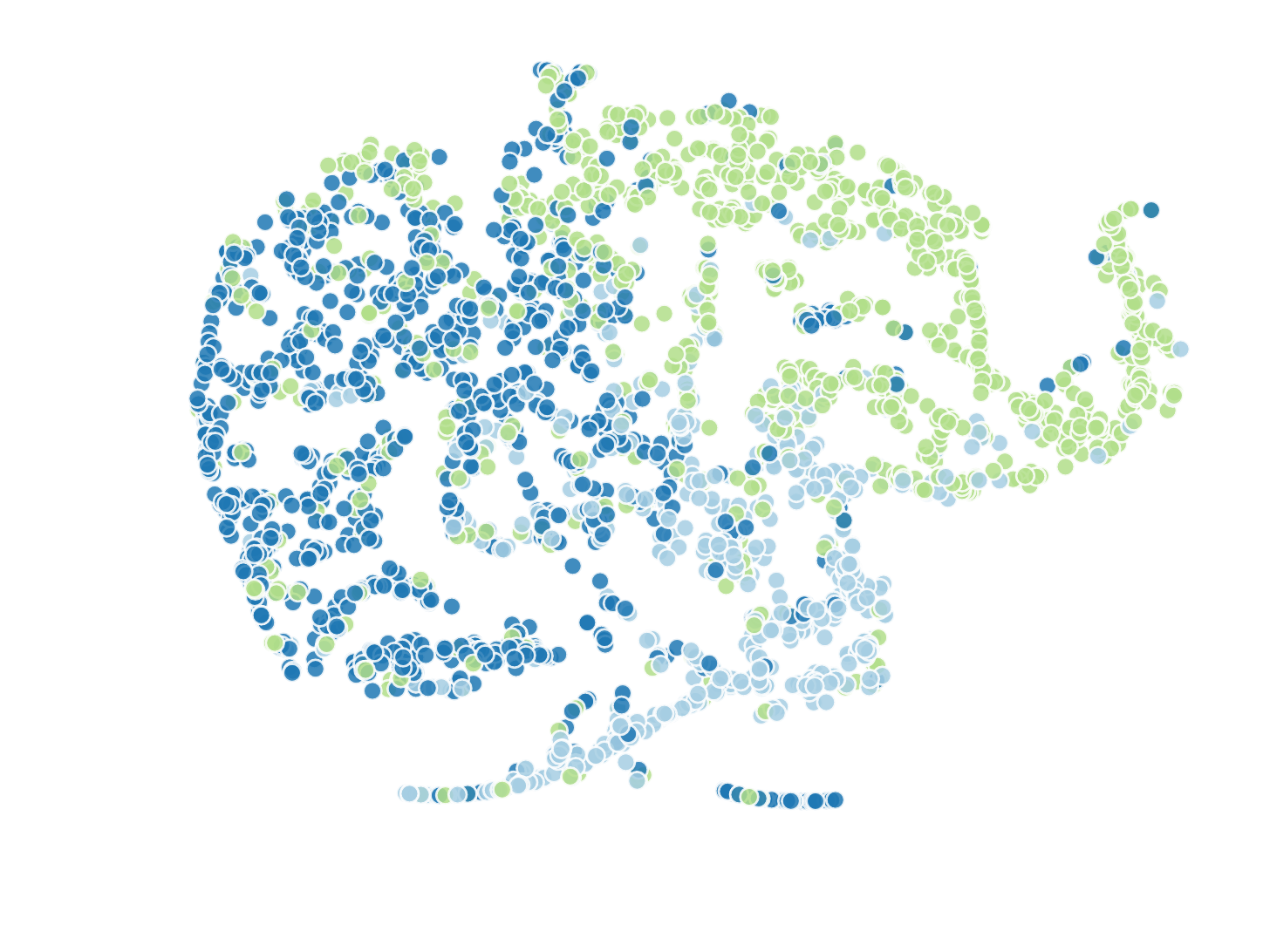}}
\subfigure[\GAT on Pubmed] {\includegraphics[width=0.245\linewidth]{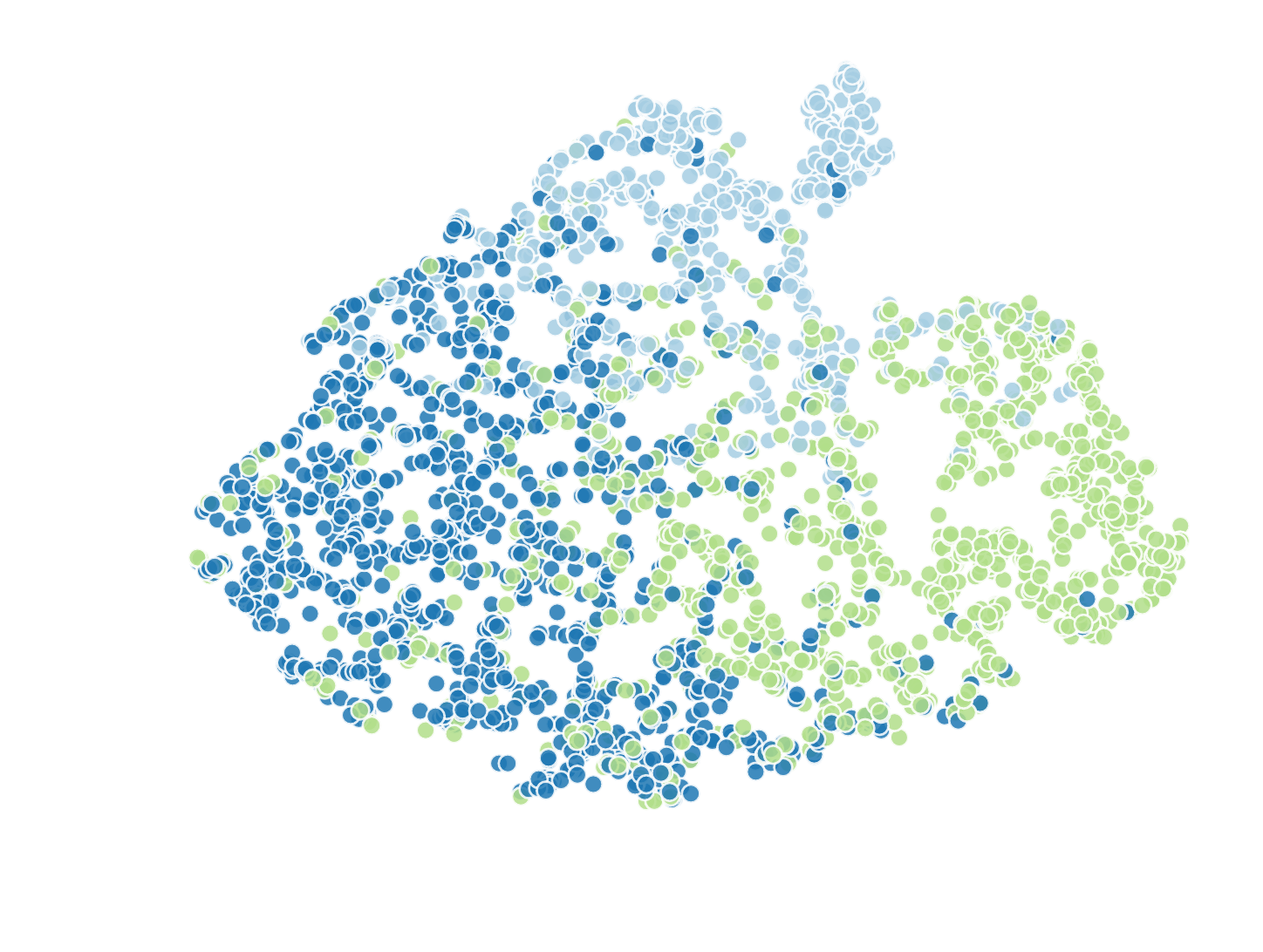}}
\subfigure [\SpGAT on Pubmed] {\includegraphics[width=0.245\linewidth]{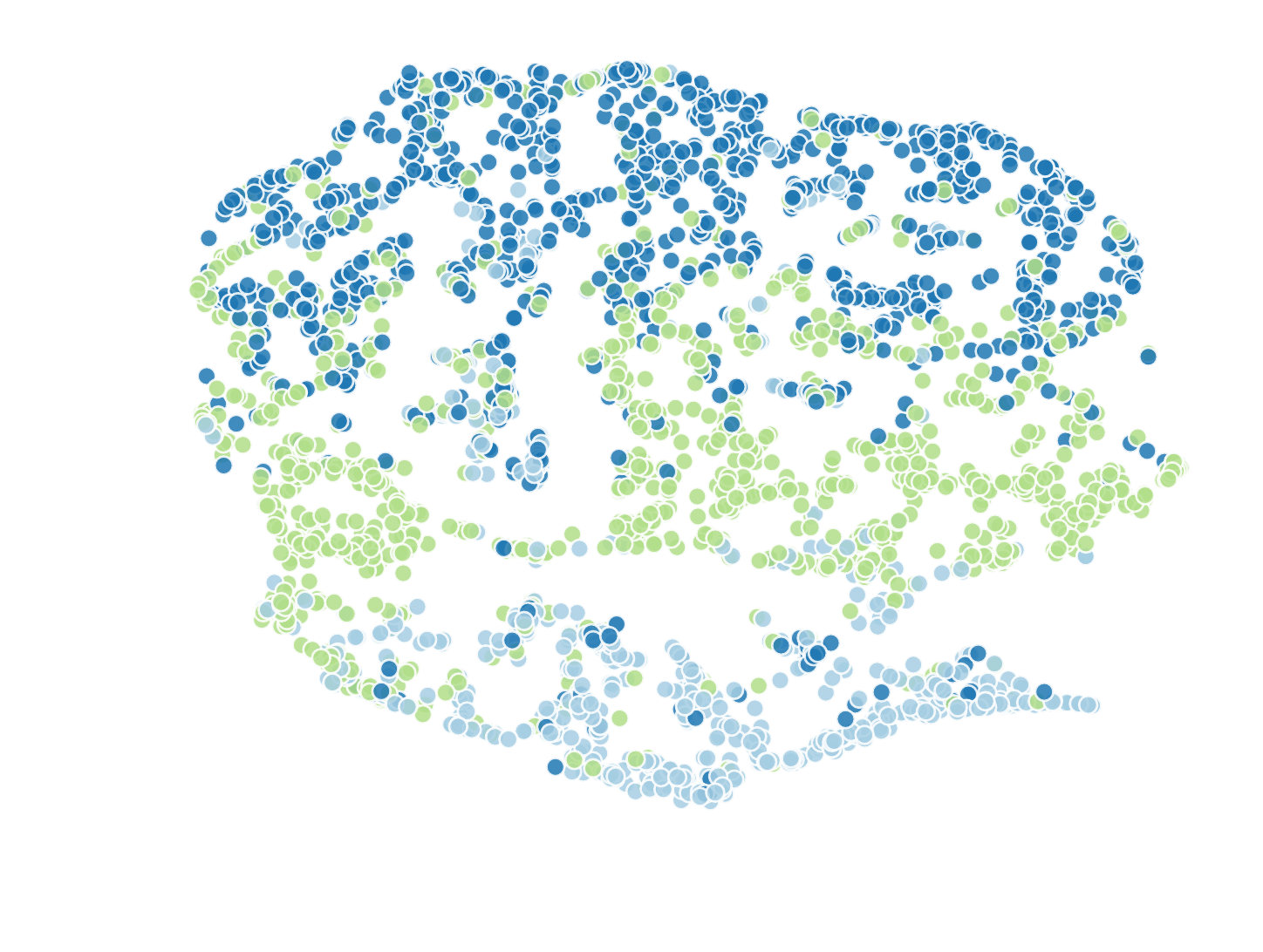}}
\vspace{-4mm}
\caption{The t-SNE visualization of \SpGAT comparing with other baselines on citation datasets. Each color corresponds to a different class that the embeddings belongs to.}
    \label{fig:tSNE}
\end{figure*}

\subsection{Simplicity of the Attention Mechanism and Choice of Spectral Bases}
To demonstrate that our proposed attention is simple but sufficient to deliver remarkably better performance, we implement a more delicate variant \SpGATNA, which applies the attention to each frequency with $\alpha=\mathrm{softmax}(\bm{a}B^{T}H\Theta^{T})$.    


Furthermore, to verify the choice of the spectral bases, we also implement a variant \SpGATF-\MAX which alters the wavelets bases to Fourier ones with \MAX-pooling.

Table~\ref{tab:simplicity} illustrates the comparison of \SpGAT with both two variants and a recent model \GCNII~\cite{chenWHDL2020gcnii}, which is a variant with many layers and has more parameters in comparison with \SpGAT-\MAX. We can find that the performances of the other two variants come worse than \SpGAT and \SpGAT achieves comparable results with \GCNII with much fewer parameters. We conjecture that involving more attention parameters could hinder the model training and bring in the risk of over-fitting, and the information contained in Fourier bases are less than wavelets, thus lead to performance detriment. However, besides the gain from using spectral wavelets as bases, the better performance that \SpGATF-\MAX has over \SpectralCNN, the version without spectral attention, also validates the effectiveness of the design of \SpGAT.

\subsection{t-SNE Visualization of Learned Embeddings}
Table~\ref{tab:Silhouette Coefficient} presents the mean Silhouette Coefficient~\cite{rousseeuw1987silhouettes} over all learned samples on three citation datasets. Larger the silhouette score is, better the clustering performs. We choose three representative baseline methods, \emph{i.e.}, \VanillaGCN~\cite{ICLR2017SemiGCN}, \GWNN~\cite{ICLR2019GWNN} and \GAT~\cite{velickovic2018gat} for comparison. We can indicate that \SpGAT achieves the best quality of embeddings.

Moreover, to evaluate the effectiveness of the learned features of \SpGAT qualitatively, we also depict the t-SNE visualization~\cite{maaten2008visualizing} of learned embeddings of \SpGAT on three citation datasets in Figure~\ref{fig:tSNE}.
The representation exhibits discernible clustering in the projected 2D space.
In Figure~\ref{fig:tSNE}, the color indicates the class label in each dataset. Compared with the other methods, the intersections of different classes in \SpGAT are more separated. It verifies the discriminative power of \SpGAT across the classes.

\subsection{Reduction of learnable parameters}

We illustrate a comparison of the number of parameters in Table~\ref{tab:parameters} on the Cora dataset with a more recent variant of GCNs, \GCNII. This reduction will be larger when the number of nodes in the graph grows. 

As supporting results, the comparable among \VanillaGCN, \GCNII, and \SpGAT on citation datasets are given as in Table~\ref{tab:comparable GCNII}. We can find that the \SpGAT achieves comparable results with \GCNII (even better on Citeseer and Pubmed) with much fewer parameters as in Table~\ref{tab:parameters}, which further indicates the effectiveness of our proposed method.

\section{Related Works}
\textbf{Spectral convolutional networks on graphs.} Existing methods of defining a convolutional operation on graphs can be broadly divided into two categories: spectral based and spatial based methods~\cite{zhang2018deep}. We focus on the spectral graph convolutions in this paper.
Spectral CNN~\cite{bruna2014spectral} first attempts to generalize CNNs to graphs based on the spectrum of the graph Laplacian and defines the convolutional kernel in the spectral domain.
ChebyNet~\cite{Defferrard2016ChebNet} introduces a fast localized convolutional filter on graphs via Chebyshev polynomial approximation.
Vanilla GCN~\cite{ICLR2017SemiGCN} further extends the spectral graph convolutions considering networks of significantly larger scale by several simplifications.
Lanczos algorithm is utilized in LanczosNet~\cite{liao2019lanczosnet} to construct low-rank approximations of the graph Laplacian for convolution.
\cite{ICLR2019GWNN} first attempts to construct graph neural networks with graph wavelets. 
SGC~\cite{wu2019simplifying} further reduces the complexity of Vanilla GCN by successively removing the non-linearities between consecutive layers. 
\cite{NIPS2019_Break} then generalizes the spectral graph convolution in block Krylov subspace forms to make use of multi-scale information.
Despite their effective performance, all these convolution theorem based methods lack the strategy to explicitly treat low- and high-frequency components with different importance.


\textbf{Space/spectrum-aware feature representation.}
In computer vision, \cite{chen2019drop} first defines space-aware feature representations based on scale-space theory and reduces spatial redundancy of vanilla CNN models by proposing the Octave Convolution (\OctConv) model.
To our knowledge, this is the first time that spectrum-aware feature representations are considered in irregular graph domain and established with graph convolutional neural networks.

\textbf{Spectral Graph Wavelets}
Theoretically, the lifting scheme is proposed for the construction of wavelets that can be adapted to irregular graphs in~\cite{sweldens1998lifting}.
\cite{Hammond2011Wavelets} defines wavelet transforms appropriate for graphs and describes a fast algorithm for computation via fast Chebyshev polynomial approximation.
For applications, \cite{tremblay2014graph} utilizes graph wavelets for multi-scale community mining and obtains a local view of the graph from each node. 
\cite{donnat2018learning} introduces the property of graph wavelets that describes information diffusion and learns structural node embeddings accordingly.
\cite{ICLR2019GWNN} first attempts to construct graph neural networks with graph wavelets. These works emphasize the local and sparse property of graph wavelets for Graph Signal Processing both theoretically and practically.

\small{
\bibliographystyle{ACM-Reference-Format}
\bibliography{www_spgat_2021}
}
\appendix